\documentclass[letterpaper,twocolumn,10pt]{article}
\usepackage{usenix-2020-09}

\usepackage{times}
\usepackage{epsfig}
\usepackage{graphicx}
\usepackage{amsmath}
\usepackage{amssymb}
\usepackage{subfigure}
\usepackage{booktabs}
\usepackage{colortbl}

\newcommand{\sname}{\mbox{{\textsc{Ex-Ray}}}\@}

\newcommand{\argmin}[1]{\underset{#1}{\operatorname{arg}\,\operatorname{min}}\;}

\newcommand{\todoc}[2]{{\textcolor{#1}{\textbf{#2}}}}

\newcommand{\todored}[1]{{\todoc{red}{\textbf{[#1]}}}}

\newcommand{\todoblue}[1]{\textcolor{blue}{#1}}

\newcommand{\xz}[1]{\todored{[XZ: #1]}}
\newcommand{\yl}[1]{\todoblue{#1}}

\makeatletter
\newcommand{\printfnsymbol}[1]{%
  \textsuperscript{\@fnsymbol{#1}}%
}
\makeatother

\date{}

\title{\sname: Distinguishing Injected Backdoor from Natural Features in Neural Networks by Examining Differential Feature Symmetry}

\author{
{\rm Yingqi Liu\thanks{equal contribution}}\\
Purdue University \\
\rm liu1751@purdue.edu
\and
{\rm Guangyu Shen\printfnsymbol{1}}\\
Purdue University \\
\rm shen447@purdue.edu
\and
{\rm Guanhong Tao}\\
Purdue University \\
\rm taog@purdue.edu
\and
{\rm Zhenting Wang}\\
Rutgers University \\
\rm zt.wang1999@gmail.com
\and
{\rm Shiqing Ma}\\
Rutgers University \\
\rm sm2283@cs.rutgers.edu
\and
{\rm Xiangyu Zhang}\\
Purdue University \\
\rm xyzhang@purdue.edu
} 

\begin{document}

\maketitle

\begin{abstract}
Backdoor attack injects malicious behavior to models such that inputs embedded with triggers are misclassified to a target label desired by the attacker. However, natural features may behave like triggers, causing misclassification once embedded.
While they are inevitable, mis-recognizing them as injected triggers causes false warnings in backdoor scanning. 
A prominent challenge 
is hence to distinguish natural features and injected backdoors. We develop a novel symmetric
feature differencing method that identifies a smallest set of features separating two classes. A backdoor is considered injected if the corresponding trigger consists of features different from the set of features distinguishing the victim and target classes. We evaluate the technique on thousands of models, including both clean and trojaned models, from the TrojAI rounds 2-4 competitions and a number of models on ImageNet. Existing backdoor scanning techniques may produce hundreds of false positives (i.e., clean models recognized as trojaned). 
Our technique removes 
78-100\%
of the false positives (by a state-of-the-art scanner ABS) with a small increase of false negatives by 0-30\%, achieving  
17-41\% overall accuracy improvement, 
and facilitates achieving top performance on the leaderboard. It also boosts performance of other scanners. 
It outperforms false positive removal methods using L2 distance and attribution techniques. 
We also demonstrate its potential in detecting a number of semantic backdoor attacks. 
\end{abstract}

\section{Introduction}
\label{sec:intro}

Backdoor attack (or trojan attack) to Deep Learning (DL) models injects malicious behaviors (e.g., by data poisoning~\cite{GuLDG19, liu2020reflection, lin2020composite, bagdasaryan2020blind} and neuron hijacking~\cite{trojannn}) such that a compromised model behaves normally for benign inputs and misclassifies to a {\em target label} when a {\em trigger} is present in the input. Depending on the form of triggers, there are 
patch backdoor~\cite{GuLDG19} where the trigger is a pixel space patch; watermark backdoor~\cite{trojannn} where the trigger is a watermark spreading over an input image, filter backdoor~\cite{liu2019abs} where the trigger is an Instagram filter, and reflection attack~\cite{liu2020reflection} 
that injects semantic trigger through reflection (like through a piece of reflective glass).
More discussion of existing backdoor attacks can be found in a few comprehensive surveys~\cite{li2020backdoor, liu2020survey, gao2020backdoor}. 
Backdoors are a prominent threat to DL applications due to the low complexity of launching such attacks, the devastating consequences especially in safety/security critical applications, and the difficulty of defense due to model uninterpretability.

There are a body of existing defense techniques.
Neural Cleanse (NC)~\cite{wang2019neural} and Artificial Brain Stimulation (ABS)~\cite{liu2019abs} make use of optimization to reverse engineer triggers and determine if a model is trojaned. Specifically for a potential target label, they use optimization to 
find a small input pattern, i.e., a trigger, that can cause any input to be classified as the target label when stamped.
DeepInspect~\cite{chen2019deepinspect} 
uses GAN to reverse engineer trigger.
These techniques leverage the observation that triggers are usually small in order to achieve stealthiness during attack such that a model is considered trojaned if small triggers can be found.
 In~\cite{jha2019attribution, erichson2020noise}, it was observed that clean and trojaned models have different behaviors under input perturbations, e.g., trojaned models being more sensitive. Such differences can be leveraged to detect backdoors.  More discussion of existing defense techniques can be found in Section~\ref{sec:related_work}.

\begin{figure}
    \centering                                          
    \footnotesize                                       
    \subfigure[Input + trigger]{
    \begin{minipage}[c]{1in}
       \center
       \includegraphics[width=01in]{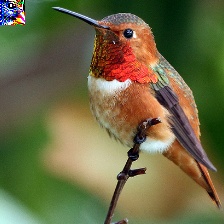}
        \end{minipage}
    }
    ~
    \subfigure[Zoomed in trigger]{
    \begin{minipage}[c]{1in}
        \center
        \includegraphics[width=1in]{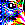}
        \end{minipage}
    }
    ~
    \subfigure[Target]{
    \begin{minipage}[c]{1in}
        \center
    
        \includegraphics[width=1in]{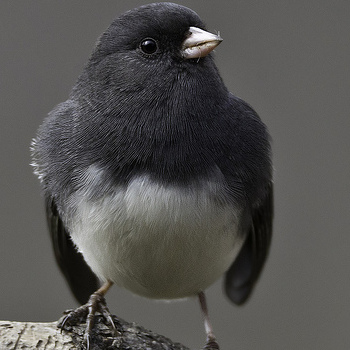}
        \end{minipage}
    }

    \caption{ImageNet model with natural trigger}

   \label{f:naturaltrigger}
\end{figure}

Although existing techniques have demonstrated effectiveness in various scenarios, an open problem is to 
distinguish natural and benign features that can act as backdoors, called {\em natural backdoors} in this paper,  from {\em injected backdoors}. As demonstrated in~\cite{liu2019abs}, natural backdoors may exist in clean models. They are usually activated using strong natural features of the target label as triggers, called {\em natural triggers} in this paper. 
Fig.~\ref{f:naturaltrigger} presents a clean ImageNet model downloaded from~\cite{Keras:online} 
with a natural trigger. Figure (a) presents a clean sample of hummingbird  stamped with a small 25$\times$25 natural trigger at the top-left corner;  (b) shows a zoom-in view of the trigger; and (c) a sample in the target class (i.e., junco bird). 
Observe that the trigger demonstrates natural features of the target (e.g., eyes and beak of junco).
The trigger can induce misclassification in 78\% 
of the hummingbird samples. 
{\em In other words, natural backdoors are due to natural differences between classes and hence inevitable}.
For instance, replacing 80\% area of any clean input with a sample of class $T$ very likely causes the model to predict $T$. 
The 80\% replaced area
can be considered a natural trigger.  
Natural backdoors share similar 
characteristics as injected backdoors, rendering the problem of distinguishing them very challenging. 
For instance, both NC and ABS rely on the assumption that triggers of injected backdoors are very small. However, natural triggers could be small (as shown by the above example) and injected triggers could be large. 
For example in semantic data poisoning~\cite{bagdasaryan2020blind}, reflection attack~\cite{liu2020reflection}, hidden-trigger attack~\cite{saha2020hidden}  and composite attack~\cite{lin2020composite}, injected triggers can be as large as any natural object. 
In the round 2 of TrojAI backdoor scanning competition\footnote{TrojAI is a multi-year and multi-round backdoor scanning competition organized by IARPA~\cite{TrojAI:online}. In each round, a large number of models of different structures are trojaned with various kinds of triggers, and mixed with clean models. Performers are supposed to identify the trojaned models. Most aforementioned techniques are being used or have been used in the competition.}, many models (up to 40\% of the 552 clean models)
have natural backdoors that are hardly distinguishable from injected backdoors, causing a large number of false positives for performers.
Examples will be discussed in Section~\ref{s:motivation}.

\begin{figure*}[htpb]
\centering
    \subfigure[Victim + \newline  injected\_trigger]{
    \begin{minipage}[t]{1in}
       \center
        \includegraphics[width=1in]{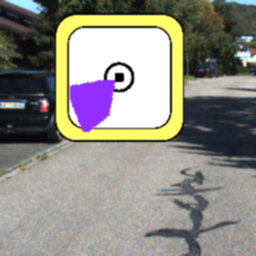}
    \end{minipage}
    }
    ~
    \subfigure[Target]{
    \begin{minipage}[t]{1in}
       \center
        \includegraphics[width=1in]{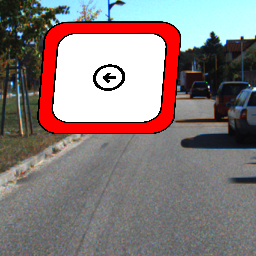}
    \end{minipage}
    }
    ~
    \subfigure[Victim+ \newline trigger\_by\_ABS]{
    \begin{minipage}[t]{1in}
       \center
        \includegraphics[width=1in]{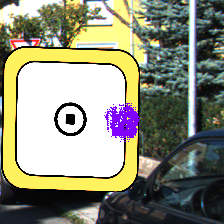}
    \end{minipage}
    }
    ~
    \subfigure[Victim]{
    \begin{minipage}[t]{1in}
       \center
       \includegraphics[width=1in]{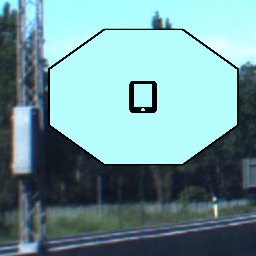}
    \end{minipage}
    }
    ~
    \subfigure[Target]{
    \begin{minipage}[t]{1in}
       \center
       \includegraphics[width=1in]{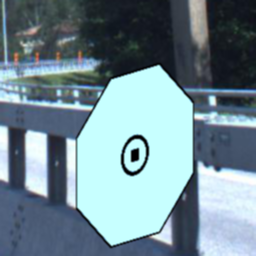}
    \end{minipage}
    }
    ~
    \subfigure[Natural trigger by ABS]{
    \begin{minipage}[b]{1in}
       \center
       \fbox{
       \includegraphics[width=0.95in]{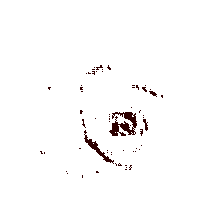}
       }
    \end{minipage}
    }
                                                                         
    \caption{
    Trojaned model \#7 from TrojAI round 2 in (a)-(c) and clean model \#123 in (d)-(f)
    }
    

   \label{f:triggerexamples}
\end{figure*}

Distinguishing natural backdoors from injected ones is critical due to the following reasons. (1) It avoids 
false warnings, which may be in a large number due to the prevalence of natural triggers. For example, any models that separate the aforementioned two kinds of birds may have natural backdoors between the two classes due to their similarity. That is, stamping any hummingbird image with  junco's beak somewhere may cause the image to be classified as junco, and vice versa. However, the models and the model creators should not be blamed for these inevitable natural backdoors. A low false warning rate is very important for traditional virus scanners. We argue that DL backdoor scanners should have the same goal. 
(2) Being correctly informed
 about the presence of backdoors and their nature (natural or injected), model end users can employ proper 
counter measures. For example,
since natural backdoors are inevitable, the users can use these models with discretion and have tolerance mechanism in place. In contrast, models with injected backdoors are just malicious and should not be used. We argue that in the future, when pre-trained models are published, some quality metrics about similarity between classes and hence natural backdoors should be released as part of the model specification to properly inform end users.
(3) Since model training is very costly, users may want to fix problematic models instead of throwing them away. Correctly distinguishing
injected and natural backdoors provides appropriate guidance in fixing. 
Note that the former denotes
out-of-distribution behaviors, which can be neutralized using in-distribution examples and/or adversarial examples. In contrast, the latter may denote in-distribution ambiguity (like cats and dogs) that can hardly be removed, or dataset biases.
For example, natural backdoors in between two classes that are not similar (in human eyes) denote that the dataset may have excessive presence of the corresponding natural features. The problem can be mitigated by improving datasets.
As we will show in Section~\ref{sec:repair_short},  injected backdoors are easier to mitigate than natural backdoors. 





We develop a novel technique to 
distinguish natural and injected backdoors. We use the following threat model.

\smallskip
\noindent
{\bf Threat Model.} Given a set of models, including both trojaned and clean models, and a small set of clean samples for each model (covering all labels), we aim to identify the models with injected backdoor(s) that can flip samples of a particular class, called the {\em victim class}, to the {\em target class}. The threat model is more general than the typical {\em universal attack} model in the literature~\cite{wang2019neural, liu2019abs, kolouri2020universal}, in which backdoors can induce misclassification for inputs of any class.  
We assume the models can be trojaned via different methods such as pixel space, feature space data poisoning (e.g., using Instagram filters as in TrojAI), composite attack, reflection attack, and hidden-trigger attack. 
According to the definition, the solution ought to prune out natural backdoor(s). $\Box$ 


We consider backdoors, regardless of natural or injected, denote differences between classes. Our overarching idea is hence to first derive a comprehensive set of natural feature differences between any pair of classes using the provided clean samples; then when a trigger is found between two classes (by an existing upstream scanner such as ABS), we compare if the feature differences denoted by the trigger is a subset of the differences between the two classes. If so, the trigger is natural. Specifically, 
let $l$ be a layer where features are well abstracted.
Given samples of any class pair, say $V$ and $T$, we aim to identify a set of neurons at $l$ such that {\em (1) if we replace the activations of the $V$ samples at those neurons with the corresponding activations of the $T$ samples, the model will classify these $V$ samples to $T$; (2) if we replace the activations of the $T$ samples at the same set of neurons with the corresponding activations of the $V$ samples, the model will classify the samples to $V$}. 
We call the conditions the {\em differential feature symmetry}.
We use optimization to identify the smallest set of neurons having the symmetry. We call it the {\em differential features} or {\em mask} in this paper. Intuitively, the features 
in the mask define the differences between the two classes.
Assume some existing backdoor scanning technique is used to generate a set of
triggers. Further assume a trigger $t$ causes
all samples in $V$ to be misclassified to $T$. We then leverage the aforementioned  
method to compute the mask between $V$ samples and $V+t$ samples, i.e., the $V$ samples stamped with $t$. Intuitively, this mask denotes the features in the trigger. 
The trigger $t$ is considered natural if its mask shares substantial commonality with the mask between $V$ and $T$.

Our contributions are summarized as follows.

\begin{itemize}
    \item We study the characteristics of natural and injected backdoors using TrojAI models and models with 
    various semantic
    backdoors. 
    \item We propose a novel symmetric feature differencing technique to distinguish the two. 

\item We implement a prototype
\sname, which stands for (``{\em \underline{\sc Ex}amining Diffe\underline{\sc r}ential Fe\underline{\sc a}ture Symmetr\underline{\sc y}}''). It can be used as an addon to serve multiple upstream backdoor scanners.

\item 
Our experiments using ABS+\sname{} on TrojAI rounds 2-4 datasets\footnote{Round 1 dataset is excluded due to its simplicity.} (each containing thousands of models), 
a few ImageNet models trojaned by data poisoning and hidden-trigger attack, 
and CIFAR10 models trojaned by composite attack 
and reflection attack, 
show that our method is highly effective in reducing false warnings (78-100\% reduction) with the cost of a small increase in false negatives (0-30\%), i.e., injected triggers are undesirably considered as natural.
It can improve multiple upstream scanners' overall accuracy including ABS (by 17-41\%), NC (by 25\%), and 
Bottom-up-Top-down backdoor scanner~\cite{sriscanner} (by 2-15\%). 
Our method also outperforms other natural backdoor pruning methods that compare L2 distances and leverage attribution/interpretation techniques. 
It allows effective detection of composite attack, hidden-trigger attack, and reflection attack that are semantics oriented (i.e., triggers are not noise-like pixel patterns but rather objects and natural features).

\item
On the TrojAI leaderboard, ABS+\sname{} achieves top performance in 2 out of the 4 rounds up to the submission day, including the most challenging round 4, with average  cross-entropy (CE) 
loss around 0.32\footnote{The smaller the better.} and average AUC-ROC\footnote{An accuracy metric used by TrojAI, the larger the better.}  around 0.90. 
It successfully beat all the round targets (for both the training sets and the test sets remotely evaluated by IARPA), which are a CE loss lower than 0.3465
More can be found in Section~\ref{sec:leaderboard}. 
\end{itemize}







\section{Motivation}
\label{s:motivation}




In this section, we use a few cases in TrojAI round 2 to study the characteristics of natural and injected backdoors and explain the challenges in distinguishing the two.  We then demonstrate our method.

According to the round 2 leader-board~\cite{TrojAI:online, pastleaderboards}, most performers cannot achieve higher than 0.80 AUC-ROC, suffering from substantial false positives. In this round, TrojAI models make use of 22 different structures such as ResNet152, Wide-ResNet101 and DenseNet201. Each model is trained to classify images of 5-25 classes. 
Clean inputs are created by compositing a foreground object, e.g.,
a synthetic traffic sign,
with a random background image from the KITTI dataset~\cite{Geiger2013IJRR}. 
An object is usually a shape with some symbol at the center. Half of the models are poisoned by stamping a polygon to foreground objects or applying an Instagram filter. 
Random transformations, such as shifting, titling, lighting, blurring, and weather effects, may be applied to improve diversity.
Fig.~\ref{f:triggerexamples} (a) to (c) show model \#7 in round 2, with (a) presenting a victim class sample 
stamped with a polygon trigger in purple, (b) a target class sample, (c) the trigger generated by ABS\footnote{Our technique requires an upstream trigger generation technique, such as ABS and NC~\cite{wang2019neural}. ABS~\cite{liu2019abs} works by optimizing a small patch in the input space that consistently flips all the samples in the victim class to the target class. It samples
internal neuron behaviors to determine the possible target labels for search space reduction. 
}.
Observe that the trigger is much smaller than the foreground objects and hence ABS can correctly determines there is a backdoor. However, there are foreground object classes similar to each other such that the features separating them could be as small as the trigger. 
Figures (d) and (e) show two different foreground object classes in model \#123 (a clean model). 
Observe that both objects are  blue octagons. The differences lie in the small symbols at the center 
of octagons. When scanning this pair of classes to determine if samples in (d) can be flipped to (e) by a trigger, ABS finds a (natural) trigger as shown 
in figure (f). Observe 
that the trigger has pixel patterns resembling the symbol at the center of (e). 
Both ABS and NC report model \#123 as trojaned (and hence a false positive) since they cannot distinguish the natural trigger from injected ones due to their similar sizes.

\begin{figure}
\centering
\vspace{0.2in}
\includegraphics[width=0.30\textwidth]{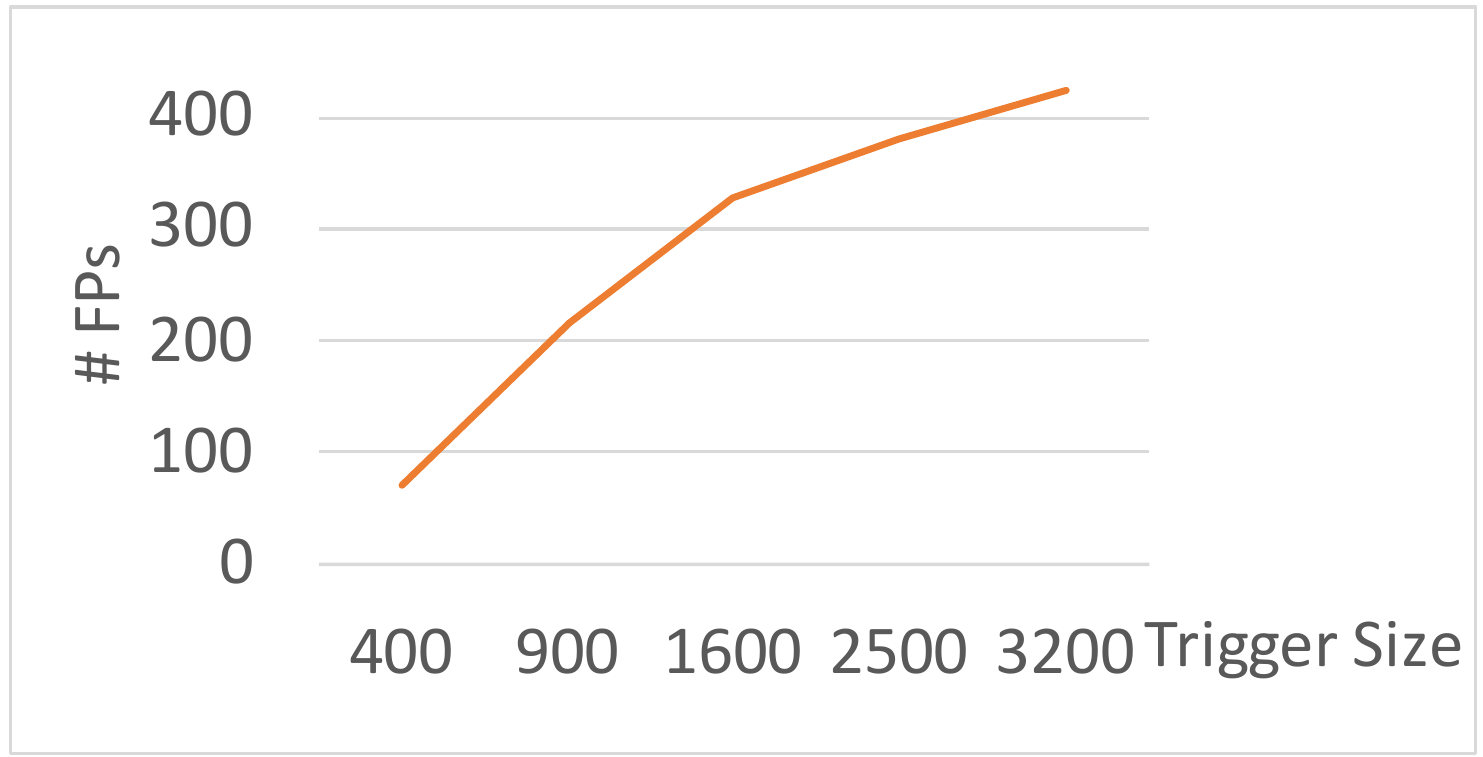}
\caption{\# FPs w.r.t trigger size}
\label{f:fp_wrt_size}
\end{figure}

Besides object classes being
too similar to each other, another reason 
for false positives in backdoor scanning is
that injected triggers could be large and complex. To detect such backdoors, optimization based techniques have to use a large size-bound in trigger reverse engineering, which unfortunately induces a lot of false positives, as large-sized natural triggers can be easily found in clean models. 
According to our analysis, the size of injected triggers in the 276 TrojAI round 2 models trojaned with polygon triggers ranges from 85 to 3200 pixels. 
Figure~\ref{f:fp_wrt_size} shows the number of false positives generated by ABS when the maximum trigger (to reverse engineer) varies from 400 to 3200 pixels. Observe that the number of false positives grows substantially with the increase of trigger size. When the size is 3200, ABS may produce over 400 false positives among the 552 clean models. 

\begin{figure}
    \centering                                                   
    \footnotesize                                                 
    \subfigure[Composite example used in poisoning]{
    \begin{minipage}[c]{0.9in}
       \center
        \includegraphics[width=0.9in]{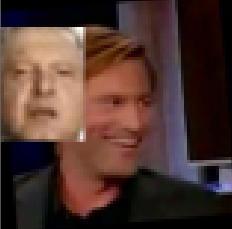}
    \end{minipage}
    }
    ~
    \subfigure[Target]{
     \begin{minipage}[c]{0.9in}
       \center
        \includegraphics[width=0.9in]{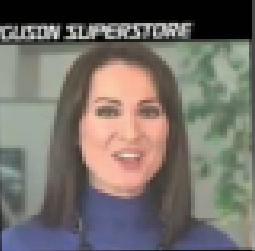}
        \end{minipage}
    }
    ~
    \subfigure[Example triggering the backdoor]{
    \begin{minipage}[c]{0.9in}
       \center
        \includegraphics[width=0.9in]{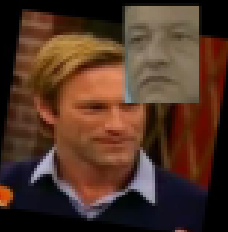}
        \end{minipage}
    }
                                                                           
    \caption{Composite attack on Youtube Face
    } 

   \label{f:compositeexamples}
\end{figure}

In addition, there are backdoor attacks that inject triggers as large as regular objects. For example, composite attack~\cite{lin2020composite} injects backdoor by mixing existing benign features from two or more classes. Fig.~\ref{f:compositeexamples} shows a composite 
attack on a face recognition model trained on the Youtube Face dataset~\cite{wolf2011face}. Figure (a) shows an  image used in the attack, mixing two persons and having the label 
set to that of (b). Note that 
the trigger is no longer a fixed 
pixel pattern, but rather the co-presence of the two persons or their features. Figure (c) shows an input that triggers the backdoor. Observe that it contains the same two persons but with different looks from (a). As shown in~\cite{lin2020composite}, neither ABS nor NC can detect such attack, as using a large trigger setting in reverse-engineering, which is needed for this scenario, produces too many false positives.

Note that although perturbation based scanning techniques~\cite{erichson2020noise, jha2019attribution} do not require optimization, they suffer from the same  problem, as indicated by the leaderboard results. This is because if benign classes are similar to each other, their classification results are as sensitive to input perturbations as classes with backdoor, causing false positives in scanning; 
if triggers are large, the injected misclassification may not be so sensitive to perturbation, causing false negatives.


\begin{figure}
    \centering                                                   
    \footnotesize     
     \subfigure[Between $V$ and $T$
     ]{
       \begin{minipage}[c]{0.70in}
       \center
       \includegraphics[width=0.73in]{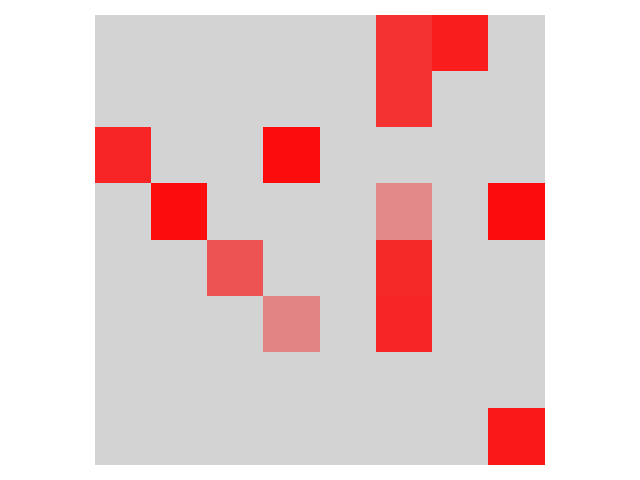}
       \end{minipage}
    }
    ~
        \subfigure[Between $V$ and $V$+trigger]{
    \begin{minipage}[c]{0.70in}
       \center
        \includegraphics[width=0.73in]{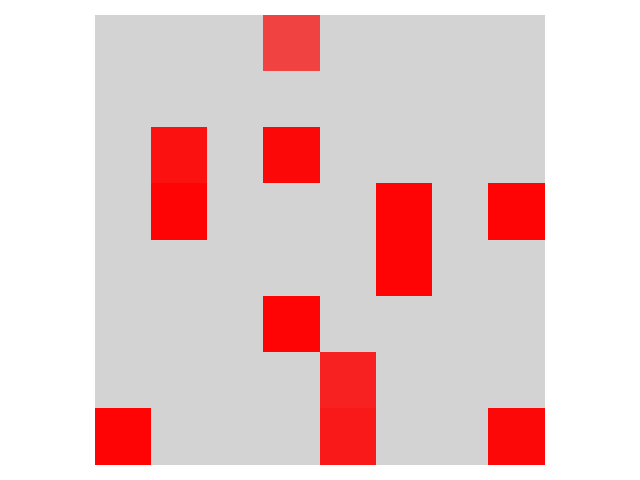}
    \end{minipage}
    }
    ~
    \subfigure[Between $V$ and $T$]{
    \begin{minipage}[c]{0.70in}
       \center
        \includegraphics[width=0.73in]{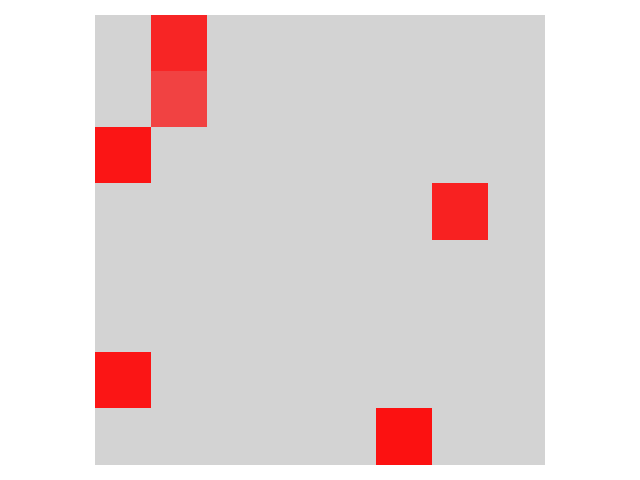}
    \end{minipage}
    }
    ~
    \subfigure[Between $V$ and $V$+trigger]{
    \begin{minipage}[c]{0.70in}
       \center
        \includegraphics[width=0.73in]{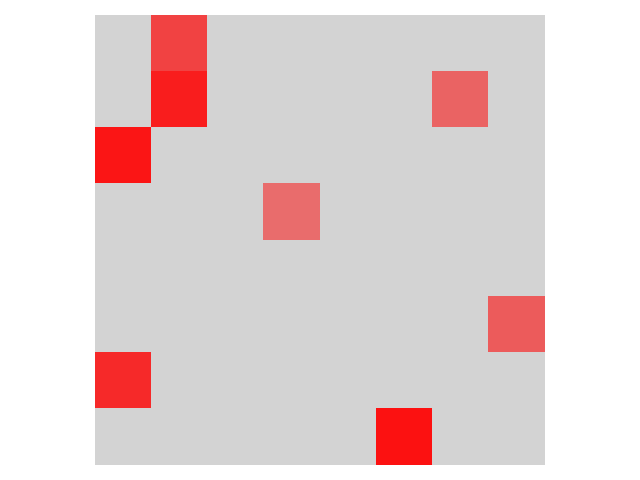}
    \end{minipage}
    }
    \caption{Differential feature masks for the trojaned model \#7 in (a)-(b) and the clean model \#123 in (c)-(d); $V$ and $T$ stand for victim and target classes, respectively.}

   \label{f:neuronmask}
\end{figure}

\smallskip
\noindent
{\bf Our Method.}
Given a trigger generated by some upstream optimization technique that flips samples in victim class $V$ to target class $T$, our technique decides if it is natural by checking if
the trigger is composed of features that distinguish $V$ and $T$. This is achieved by a symmetric feature differencing method. The method identifies a set of features/neurons (called mask) such that copying their values from one class to another flips the classification results. 
Fig.~\ref{f:neuronmask} (a) shows the mask in the second last convolutional layer of the TrojAI round 2 trojaned model \#7 (i.e., the model in Fig.~\ref{f:triggerexamples} (a)-(c)). It distinguishes the victim and target classes. Note that {\em a mask is not specific to some input sample, but rather to an pair of classes}. 
Each block in the mask denotes a feature map (or a neuron) with 
red meaning that the whole feature map needs to be copied (in order to flip classification results); gray meaning that the map is not necessary; and light red meaning that the copied values are mixed with the original values. 
As such, copying/mixing the activation values of the red/light-red neurons from the target class samples to the victim class samples can flip all the victim samples to the target class, and vice versa.
Intuitively, it denotes the features distinguishing the victim and target classes.
Fig.~\ref{f:neuronmask} (b) shows the mask that distinguishes the victim class samples and the victim samples stamped with the trigger, denoting the constituent features of the trigger.
Observe that (a) and (b) do not have a lot in common.
Intuitively, the trigger consists of many features that are not part of the distinguishing features of the victim and target classes. In contrast, Fig.~\ref{f:neuronmask} (c) and (d) show the corresponding masks for the clean model \#123 in Fig.~\ref{f:triggerexamples} (d)-(f). Observe that (c) and (d) have a lot in common, indicating the trigger consists of most the features distinguishing the victim and target classes and is hence natural.

\begin{figure}
    \centering                                                   
    \footnotesize
    \subfigure[Target]{
    \begin{minipage}[c]{0.6in}
       \center
        \includegraphics[width=0.8in]{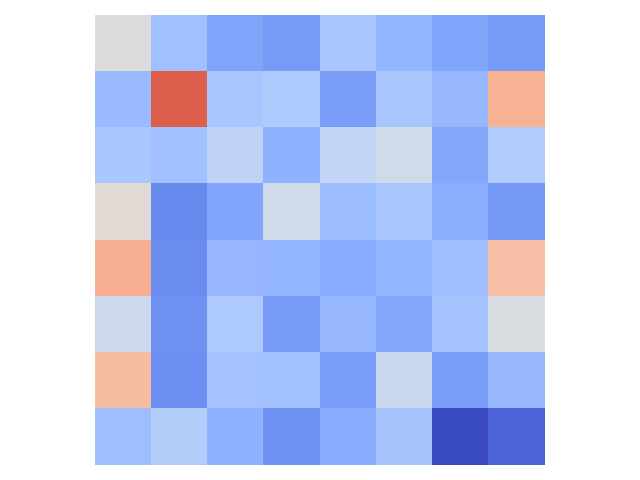}
    \end{minipage}
    }                              
    ~
    \subfigure[Victim+trigger (L2: 0.59)]{
    \begin{minipage}[c]{0.8in}
       \center
        \includegraphics[width=0.8in]{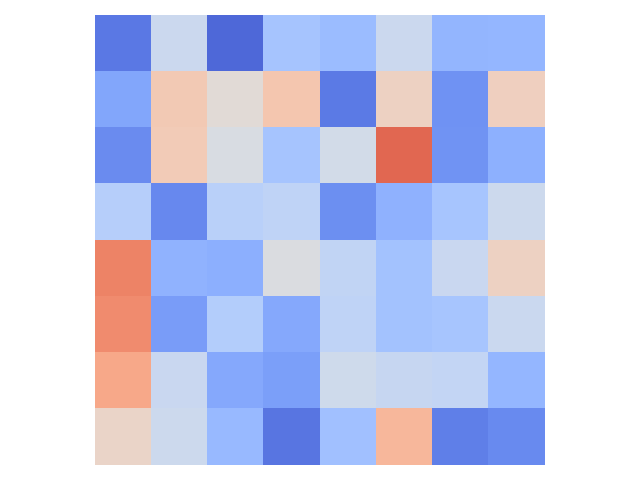}
    \end{minipage}
    }  
    ~
    \subfigure[Target]{
    \begin{minipage}[c]{0.6in}
       \center
        \includegraphics[width=0.9in]{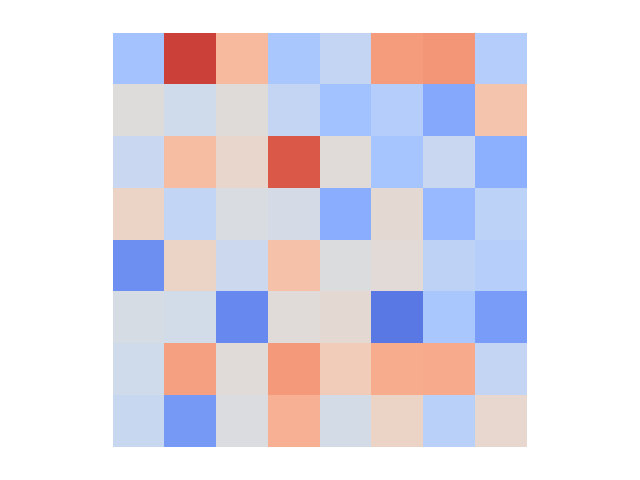}
    \end{minipage}
    }
    ~
    \subfigure[Victim+trigger (L2: 0.74)]{
    \begin{minipage}[c]{0.8in}
       \center
        \includegraphics[width=1in]{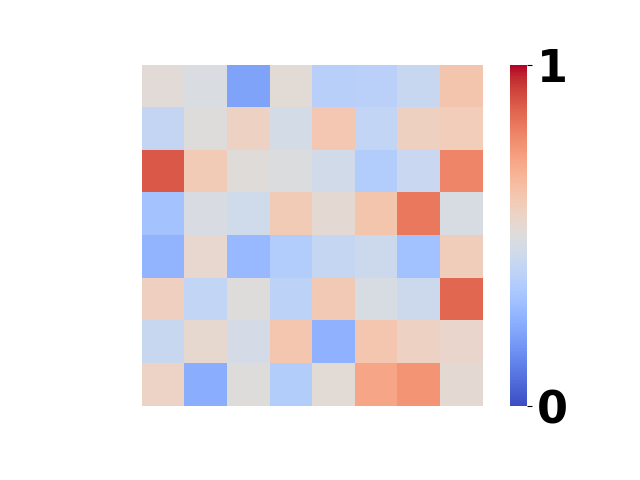}
    \end{minipage}
    }
                             
    \caption{ Feature maps of the trojaned model \#7 in (a)-(b) and the clean model \#123 in (c)-(d). Each color block denotes the average value of a feature map after normalization 
    } 
    
   \label{f:featuremap}
   \vspace{-0.2in}
\end{figure}

Directly comparing activation values does not work. Fig.~\ref{f:featuremap} (a) and (b) show the average feature maps at the second last convolutional layer for the target class samples and the victim class samples stamped with the trigger, respectively, for the aforementioned trojaned model \#7.
Each block denotes the mean of the normalized activation values in a feature map.
The L2 distance of the two is 0.59 as shown in the caption of (b). 
It measures the distance between the stamped samples and the clean target class samples. Ideally, a  natural trigger would yield a small L2 distance as it possesses the target class features.
Figures (c) and (d) show the corresponding information for a clean model, with the L2 distance 0.74 (larger than 0.59). Observe there is not a straightforward separation of the two. This is because such a simple method does not consider what features are critical. Empirical results can
be found in Section~\ref{sec:eval}.

\begin{figure}
    \centering                                                   
    \footnotesize
    \subfigure[Target]{
    \begin{minipage}[c]{0.7in}
       \center
        \includegraphics[width=0.8in]{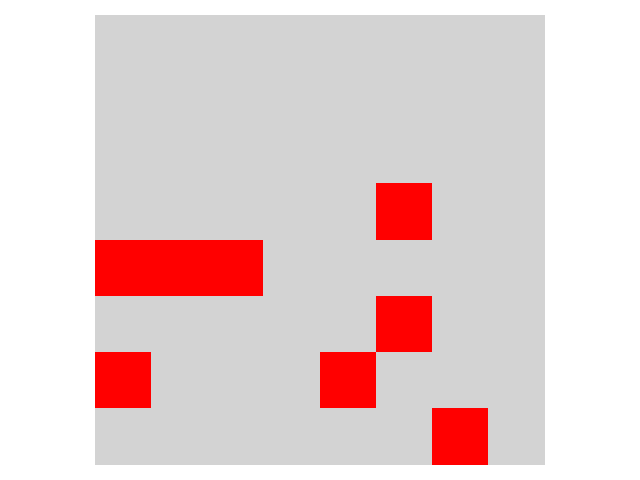}
    \end{minipage}
    }                              
    ~
    \subfigure[Victim+ \newline trigger (L2:0.22)]{
    \begin{minipage}[c]{0.7in}
       \center
        \includegraphics[width=0.8in]{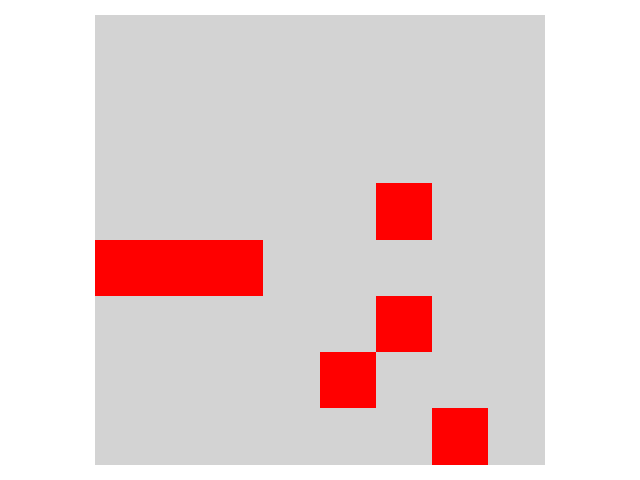}
    \end{minipage}
    }  
    ~
    \subfigure[Target]{
    \begin{minipage}[c]{0.6in}
       \center
        \includegraphics[width=0.8in]{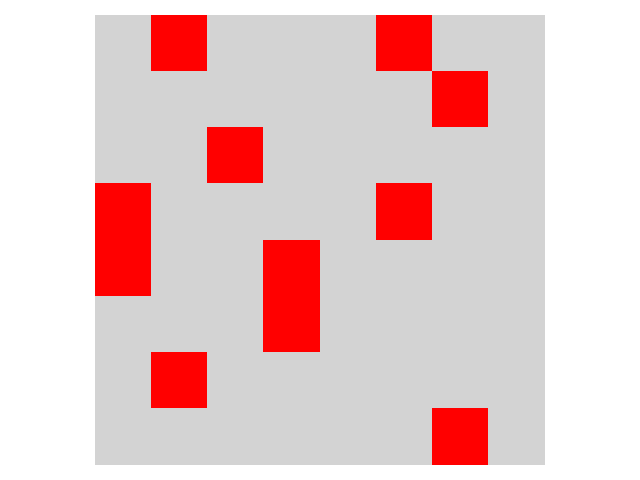}
    \end{minipage}
    }
    ~
    \subfigure[Victim+\newline trigger (L2:0.21)]{
    \begin{minipage}[c]{0.8in}
       \center
        \includegraphics[width=0.8in]{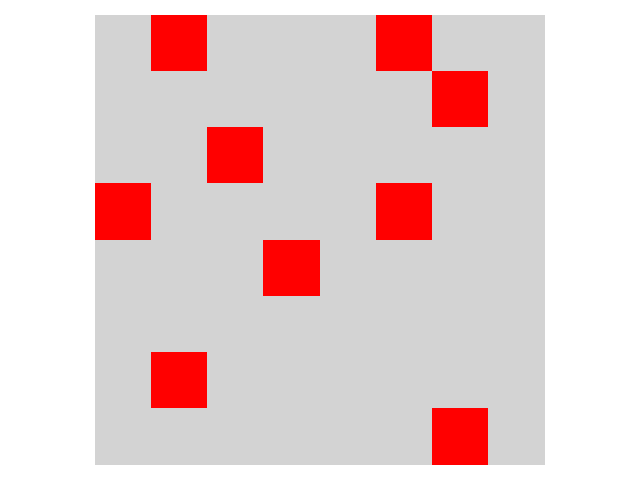}
    \end{minipage}
    }
                             
    \caption{Important neurons by an attribution technique with (a)-(b) for model \#7 and (c)-(d) for model \#123} 

   \label{f:attributemap}
\end{figure}

A plausible improvement is to use attribution techniques to identify the important neuron/features and only compare their values.
Fig.~\ref{f:attributemap} (a) and (b) present the 10\% most important neurons in the trojaned model \#7 identified by an attribution technique DeepLift~\cite{shrikumar2016not}, for the target class samples and victim samples with the trigger, respectively.  
The L2 distance for the features in the intersection of the two (i.e., the features important in both) is 0.22 as shown in the caption of (b). 
Figures (c) and (d) show the information for the clean model \#123 (with L2 distance 0.21). Even with the attribution method, the two are not that separable.
This is because these techniques  identify {\em features that are important for a class or a sample} whereas our technique identifies {\em comparative importance}, i.e., {\em features that are important to distinguish two classes}.
The two are quite different as shown by the differences between Fig.~\ref{f:neuronmask} and Fig.~\ref{f:attributemap}. 
More results can be found in Section~\ref{sec:eval}. 

Applying a symmetric differential analysis similar to ours in the input space does not work well either. This is because 
semantic features/objects may appear in different positions of the inputs. Differencing pixels without aligning corresponding features/objects likely yields meaningless results.



\section{Design}
\label{sec:design}

\begin{figure}
    \centering                                                   
    \footnotesize 

        \includegraphics[width=0.46\textwidth]{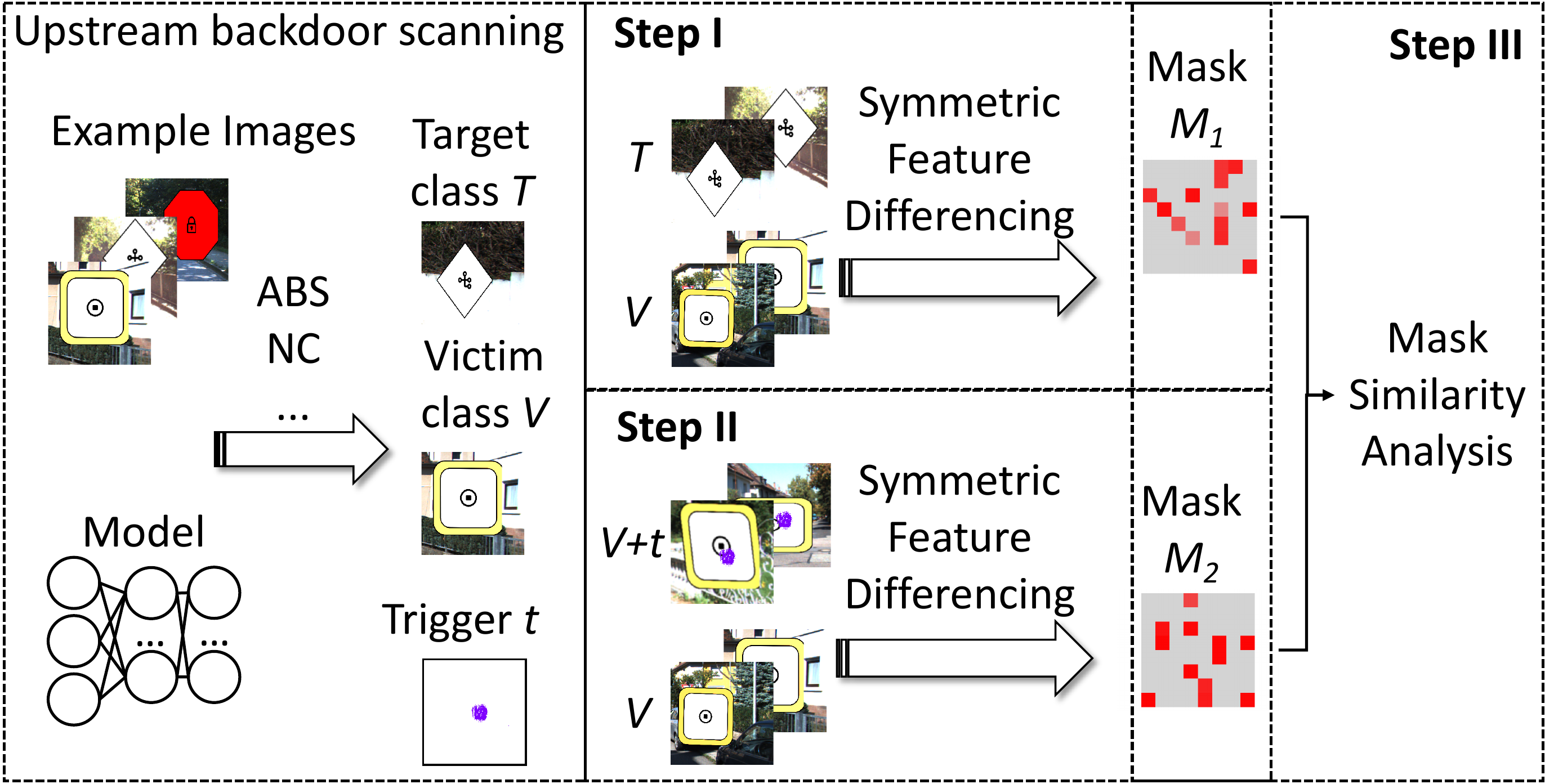}

    \caption{\sname{} workflow}
    \vspace{-0.2in}

   \label{f:overview}
\end{figure}

Fig.~\ref{f:overview} shows the overview of \sname. The key component is the symmetric featuring differencing analysis. 
Given two sets of inputs, it 
identifies features of {\em comparative importance}, i.e., distinguishing the two sets. We also call them {\em differential features} or a {\em mask}
for simplicity. 
Given a trigger $t$ generated by some upstream scanning technique (not our contribution)  that flips class $V$ samples to class $T$. In step I, the technique
computes the mask separating $V$ and $T$.
In step II, it computes the mask separating $V$ and $V$+$t$ samples, which are classified to $T$. In step III, a similarity analysis is used to compare the two masks to determine if the trigger 
is natural. Next we will explain the components in details. 


\subsection{Symmetric Feature Differencing}
\label{s:sfd}
Given two sets of inputs of classes $V$ and $T$, respectively, and a hidden layer $l$,
the analysis identifies a smallest set of features/neurons $M$ (or mask) such that if 
we copy the values of $M$ from $T$ samples to $V$ samples, we can flip the $V$ samples to class $T$, and vice versa.
Note that such a set must exist as in the worse case, it is just the entire set of features at $l$. We hence use optimization to find the smallest set.

\begin{figure*}[]
    \centering
    \includegraphics[width=0.8\textwidth]{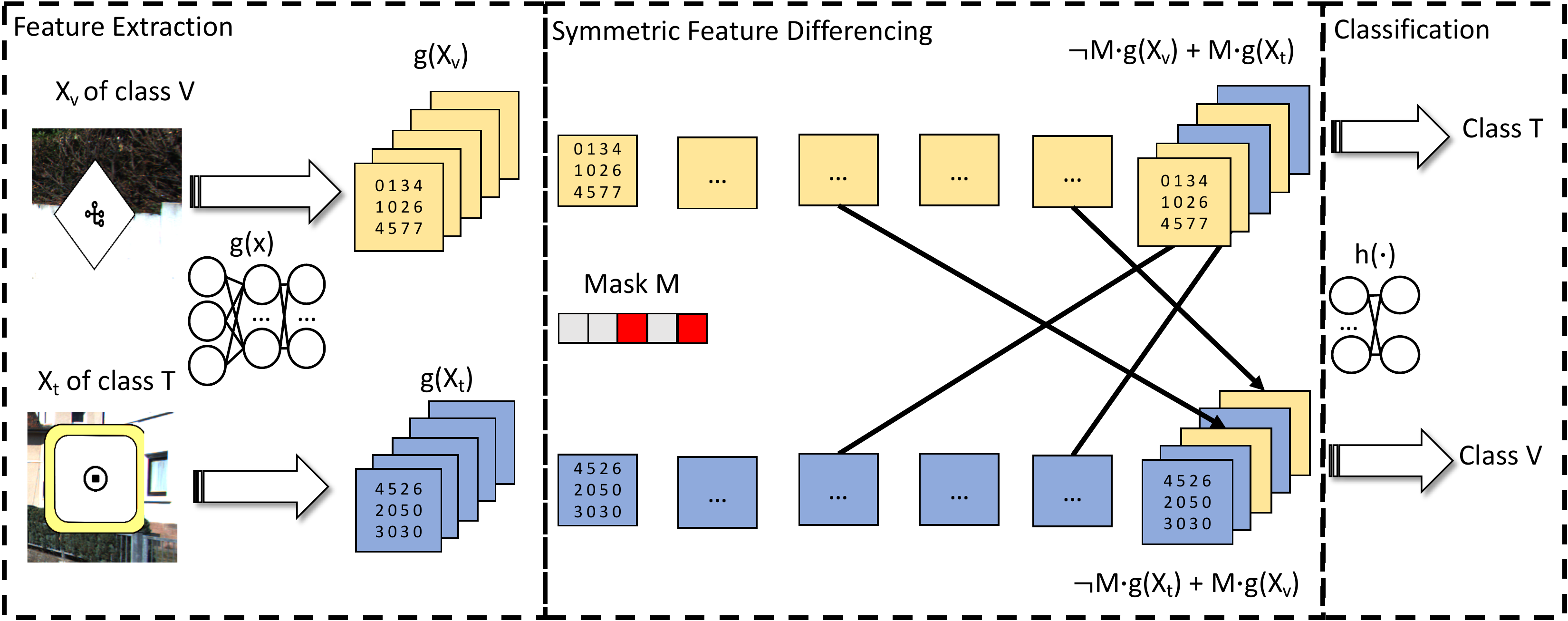}
    \caption{Example to Illustrate Symmetric Feature Differencing}
  \label{f:maskexample}
\end{figure*}

\smallskip
\noindent
{\bf Differencing Two Inputs.}
For explanation simplicity, we first discuss how the technique works on two inputs: $x_v$ in $V$ and $x_t$ in $T$. We then expand it to compare two sets. 

Let $F(x)$ be a feed forward neural network under analysis. Given the inner layer $l$, let $g$ be the sub-model up to layer $l$ and $h$ the sub-model after $l$.
Thus, $F(x) = h(g(x))$. Let the number of features/neurons at $l$ be $n$. The set of differential features (or mask) $M$ is denoted as an $n$ element vector with values in $[0,1]$. $M[i]=0$ means that the $i$th neuron is not a differential feature; $M[i]=1$ means $i$ is a {\em must-differential} feature; and $M[i]\in (0,1)$ means $i$ is a {\em may-differential} feature. The must and may features function differently during value copying, which will be illustrated later.
Let $\neg M$ be the negation of the 
mask such that $\neg M[i]=1-M[i]$ with $i\in [1,n]$.
The aforementioned symmetric property is hence denoted as follows. 
\begin{equation}\label{e:cns1}
h(g(x_v)\cdot M  + g(x_t)\cdot \neg M) = V
\end{equation}
\begin{equation}\label{e:cns2}
h(g(x_v)\cdot \neg M  + g(x_t)\cdot  M) = T
\end{equation}
Intuitively, when $M[i]=0$, the original $i$th feature map is retained; when $M[i]=1$, the $i$th feature map is replaced with that from the other sample; when $M[i]\in (0,1)$, the feature map is the weighted sum of the $i$th feature maps of the two samples.
{\em The reasoning is symmetric because the differences between two samples is a symmetric relation.}

\smallskip
\noindent
\underline{\em Example. } 
Figure~\ref{f:maskexample} illustrates an example of symmetric feature differencing.  The box on the left shows the $g(x)$ function and that on the right the $h(\cdot)$ function. 
The top row in the left box shows that
five feature maps (in yellow) are generated by $g()$
for a victim class sample $x_v$.
The bottom row shows that
the five feature maps (in blue) 
for a target class sample $x_t$.
The box in the middle illustrates the 
symmetric differencing process. 
As suggested by the red entries in the mask $M$ in the middle (i.e., $M[3]=M[5]=1$), 
in the top row, the 3rd and 5th (yellow) feature maps are replaced with the corresponding (blue) feature maps from the bottom. Symmetric replacements happen in the bottom row as well. 
On the right, the mutated feature maps flip the classification results to $T$ and to $V$, respectively. $\Box$


\smallskip
\noindent{\bf Generating Minimal Mask by Optimization.}
The optimization to find the minimal $M$ is hence the following.
\begin{align}\label{e:cnm1}
\begin{split}
& \argmin{M}{sum(M)}, s.t. \\
& h(g(x_v)\cdot M  + g(x_t)\cdot \neg M) = V \ \mathit{and} \\
& h(g(x_v)\cdot \neg M  + g(x_t)\cdot  M) = T
\end{split}
\end{align}
To solve this optimization problem, we devise a loss in~(\ref{e:cnmloss}). It has three parts. The first part $sum(M)$ is to minimize the mask size. 
The second part $w_1\times ce_1$ is a barrier loss for constraint~(\ref{e:cns1}), with $ce_1$ the cross entropy loss when replacing $x_t$'s features.
Coefficients $w_1$ is dynamic. When the cross entropy loss is larger than 
a threshold $\alpha$, 
$w_1$ is set to a large value $w_{large}$. This forces $M$ to satisfy constraint~(\ref{e:cns1}).
When the loss is small indicating the constraint is satisfied, $w_1$ 
is changed to a small value $w_{small}$. 
The optimization hence focuses on minimizing the mask.
The third part $w_2\times ce_2$ is similar.
\begin{align}\label{e:cnmloss}
\begin{split}
\mathcal{L}_{pair}& (x_v, x_t) = sum(M) + w_1 \times ce_1 + w_2 \times ce_2,\  \\
\mathit{with}\ & ce_1 = \mathit{CE}(h( g(x_v)\cdot M + g(x_t)\cdot \neg M), V ),  \\
& ce_2 = \mathit{CE}(h( g(x_v)\cdot \neg M + g(x_t)\cdot M), T ),  \\
& w_1 = w_{large} \text{ if } ce_1 > \alpha \text{ else } w_{small},   \\
& w_2 = w_{large} \text{ if } ce2 > \alpha \text{ else } w_{small}
\end{split}
\end{align}

\noindent
{\bf Differencing Two Sets.}
The algorithm to identify the differential features of two sets can be built from the primitive of comparing two inputs.
Given two sets $X_V$ of class $V$ and $X_T$ of class $T$, ideally the mask $M$ should satisfy the constraints (\ref{e:cns1}) and (\ref{e:cns2}) for any $x_v\in X_V$ and $x_t\in X_T$.
While such a mask must exist (with the worst case containing all the features), minimizing it becomes very costly. Assume $|X_V|=|X_T|=m$. The number of constraints that need to be satisfied during optimization is $O(m^2)$.
Therefore, we develop a stochastic method that is $O(m)$. Specifically, let $\overrightarrow{X_V}$ and $\overrightarrow{X_T}$ be random orders of $X_V$ and $X_T$, respectively. We minimize $M$ such that
it satisfies constraints (\ref{e:cns1}) and (\ref{e:cns2}) for all pairs ($\overrightarrow{X_V}[j]$, $\overrightarrow{X_V}[j]$), with $j\in [1,m]$.
Intuitively, we optimize on a set of random pairs from $X_V$ and $X_T$ that cover all the individual samples in $X_V$ and $X_T$. 
The loss function is hence the following.
\[\mathcal{L}=\sum\limits_{j=1}^{m}\mathcal{L}_{pair}(\overrightarrow{X_V}[j],\ \overrightarrow{X_T}[j])\]
When $X_V$ and $X_T$ have one-to-one mapping, such as the victim class samples and their compromised versions that have the trigger embedded, we can directly use the mapping in optimization instead of a random mapping.
We use Adam optimizer~\cite{kingma2014adam} with a learning rate 5e-2 and 400 epochs. Masks are initialized to all 1 to begin with. This denotes a conservative start since such masks suggest swapping all feature maps, which must induce the intended classification results swap. 

\begin{figure}[htbp]
    \centering                                                   
    \footnotesize
    \subfigure[T]{
    \begin{minipage}[c]{0.55in}
       \center
        \includegraphics[width=0.5in]{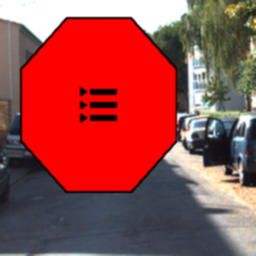}
    \end{minipage}
    }                              
    ~
    \subfigure[0]{
    \begin{minipage}[c]{0.55in}
       \center
        \includegraphics[width=0.7in]{figs/id-00000004_gradual_2_hm0_0.png}
    \end{minipage}
    }                              
    ~
    \subfigure[30]{
    \begin{minipage}[c]{0.55in}
       \center
        \includegraphics[width=0.7in]{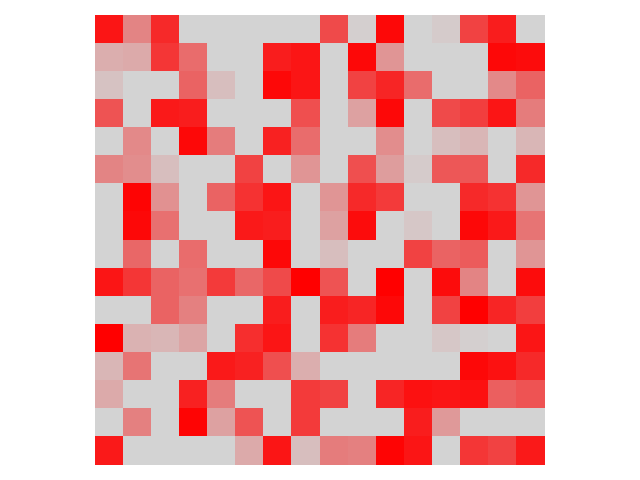}
    \end{minipage}
    }  
    ~
    \subfigure[60]{
    \begin{minipage}[c]{0.55in}
       \center
        \includegraphics[width=0.7in]{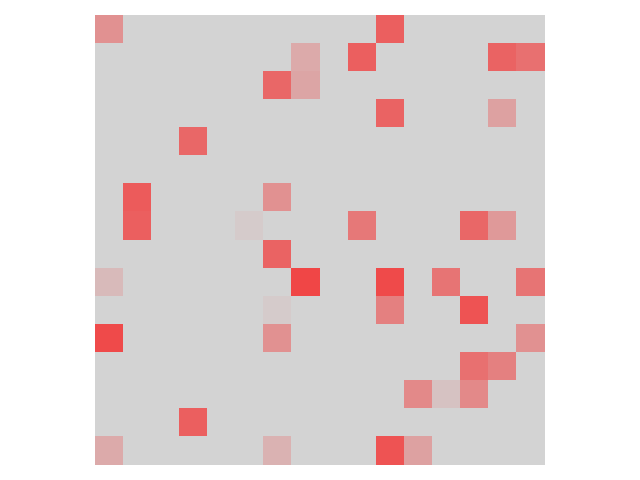}
    \end{minipage}
    }
    ~
    \subfigure[90]{
    \begin{minipage}[c]{0.55in}
       \center
        \includegraphics[width=0.7in]{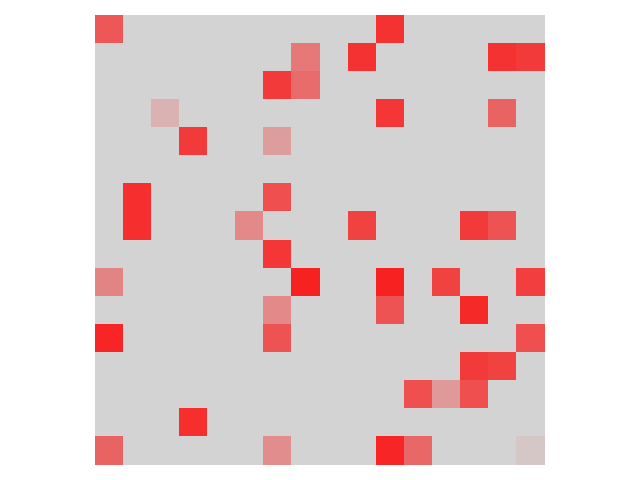}
    \end{minipage}
    }
    \\
    \subfigure[V+trigger]{
    \begin{minipage}[c]{0.55in}
       \center
        \includegraphics[width=0.5in]{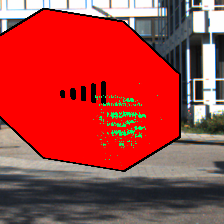}
    \end{minipage}
    }                              
    ~
    \subfigure[0]{
    \begin{minipage}[c]{0.55in}
       \center
        \includegraphics[width=0.7in]{figs/id-00000004_gradual_2_hm1_0.png}
    \end{minipage}
    }                              
    ~
    \subfigure[30]{
    \begin{minipage}[c]{0.55in}
       \center
        \includegraphics[width=0.7in]{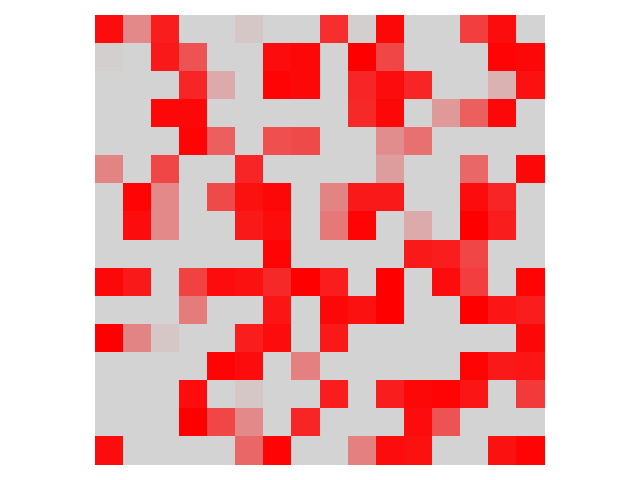}
    \end{minipage}
    }  
    ~
    \subfigure[60]{
    \begin{minipage}[c]{0.55in}
       \center
        \includegraphics[width=0.7in]{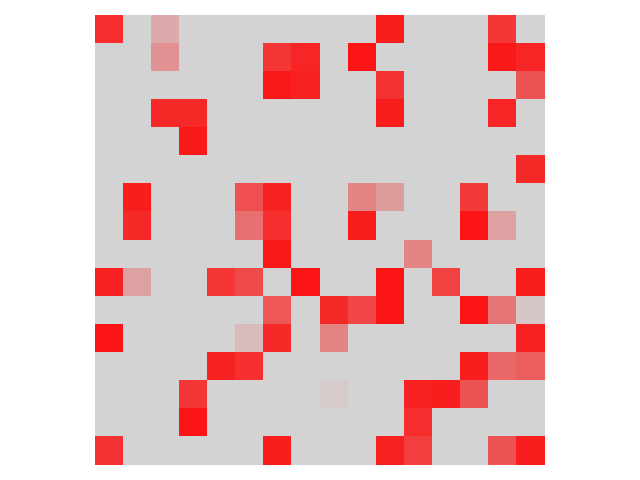}
    \end{minipage}
    }
    ~
    \subfigure[90]{
    \begin{minipage}[c]{0.55in}
       \center
        \includegraphics[width=0.7in]{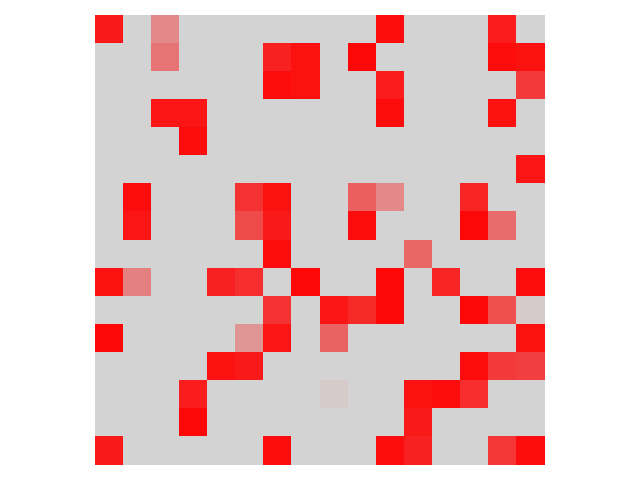}
    \end{minipage}
    }
                     
    \caption{Mask changes over time related to classes \# 2 ($V$) and \#3 ($T$) in a benign model \#4 in TrojAI round 2, with (f) a $V$ sample containing a trigger generated by ABS (the small patch close to the central symbol) classified as $T$; (b)-(e) show the mask between $V$ and $V$+trigger 
    after different numbers of optimization epochs; (g)-(j) the mask between $V$ and $T$.
    }
   \label{f:maskchange}
\end{figure}

\smallskip
\noindent
\underline{\em Example.}
Figure~\ref{f:maskchange} shows how masks change over time for a benign model \#4 from TrojAI round 2. Figure (a) shows the target class $T=\#3$ and (f) the victim class $V=\#2$ with a trigger generated by ABS (close to the center resembling the symbol in the middle of target class).
Observe that the two classes are similar and hence ABS generates a small trigger that can flip $V$ to $T$. Figures (b)-(e) show the changes of mask between $V$ and $V$+trigger and (g)-(j) for the mask between $V$ and $T$. Observe that 
in both cases, the initially all-1 masks (i.e., all red) are reduced to having sparse 1's and some smaller-than-1 values. The masks in the top row consistently share substantial commonality with  those in the bottom row, suggesting the similarity of the differential features. $\Box$

\smallskip
\noindent
{\bf Symmetry Is Necessary.}
Observe that our technique is symmetric. Such symmetry is critical to effectiveness. 
One may wonder that a one-sided analysis that only enforces constraint (\ref{e:cns2}) may be sufficient. That is, $M$ is the minimal set of features that when copied from $T$ (target) samples to $V$ (victim) samples can flip the $V$ samples to class $T$. However, this is insufficient. In many cases, misclassfication (of a $V$ sample to $T$) can be induced when strong features of class $V$ are suppressed (instead of adding strong $T$ features). Such $V$ features cannot be identified by the aforementioned one-sided analysis, while they can be identified by the analysis along the opposite direction (i.e., constraint (\ref{e:cns1})). Our experiments in Section~\ref{sec:eval} show the importance of symmetry.

\begin{figure}[htbp]
    \centering                       
    \footnotesize

    \subfigure[V]{
    \begin{minipage}[c]{0.52in}
       \center
        \includegraphics[width=0.52in]{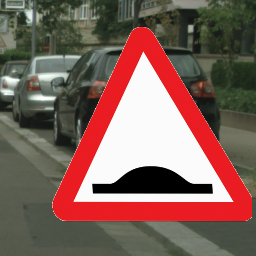}
    \end{minipage}
    }                    ~
    \subfigure[T]{
    \begin{minipage}[c]{0.52in}
       \center
        \includegraphics[width=0.52in]{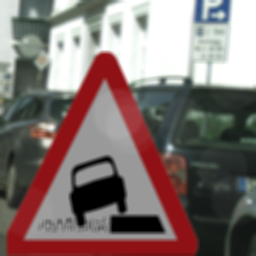}
    \end{minipage}
    }  
    ~
    \subfigure[V+trigger]{
    \begin{minipage}[c]{0.55in}
       \center
        \includegraphics[width=0.52in]{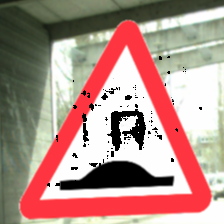}
    \end{minipage}
    }
    ~
    \subfigure[1 sided mask from V to T]{
    \begin{minipage}[c]{0.55in}
       \center
        \includegraphics[width=0.7in]{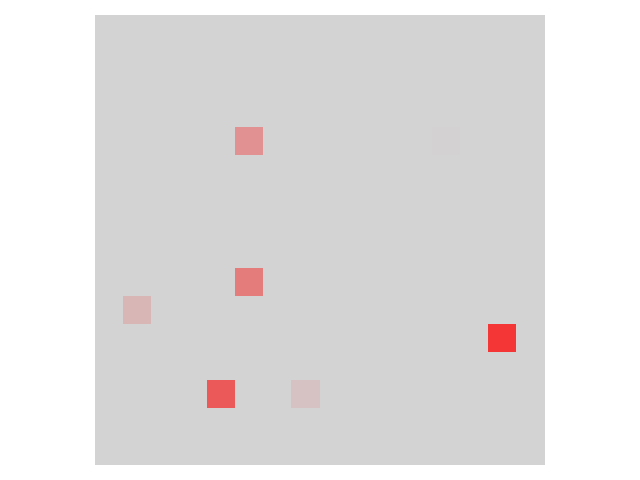}
    \end{minipage}
    }                                 ~
    \subfigure[1 sided mask from V to V+trigger]{
    \begin{minipage}[c]{0.55in}
       \center
        \includegraphics[width=0.7in]{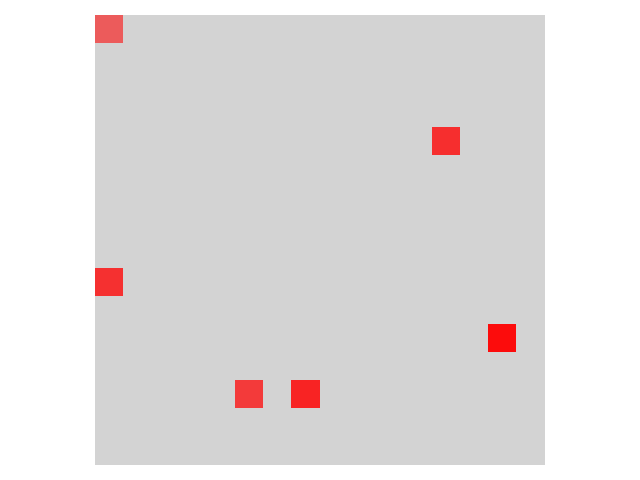}
    \end{minipage}
    }  
     
    \caption{One sided masks for model \# 18 in round 3 with victim class $V=$\#8 and target class $T=$\#3}
   \label{f:symmetry}
\end{figure}

\smallskip
\noindent
\underline{\em Example.}
Fig.~\ref{f:symmetry} presents an example for one-sided masks
from a clean model \#18 in TrojAI round 3. 
Figures (a) and (b) present the victim and target classes and (c) a natural trigger generated by ABS, which resembles the central symbol in the 
target class. Figure (d) shows the
one-sided mask from $V$ to $T$,
meaning that copying/mixing the feature maps as indicated by the mask from $T$ samples 
to $V$ samples can flip the classification results to $T$.
Figure (e) shows the one-sided mask from $V$ to $V$+trigger. 
Note that $V$+trigger samples are classified to $T$. 
Although in both cases $V$
samples are flipped to $T$, the
two one-sided masks have only one entry in common, suggesting that the ways they induce the classification results are different. 
In contrast, the symmetric masks (not shown due to space limitations) share a lot of commonality. $\Box$

\subsection{Comparing Differential Feature Sets To Identify Natural Backdoors}
\label{s:similarity}
As shown in Fig.~\ref{f:overview}, we 
first compute the differential features  between the victim and target classes, denoted as
$M_1$, and then those between the victim samples and their compromised versions,
denoted as $M_2$. Next, we explain how to compare $M_1$ and $M_2$ to determine if the trigger is natural.
Intuitively $M_1$ and $M_2$ should share a lot of commonality when the trigger is natural, as reflected in the following condition.
\begin{align}\label{e:fprule1}
\small
\begin{split}
&    sum(min(M_1, M_2))   > \beta \times min(sum(M_1), sum(M_2))
\end{split}
\end{align}
In (\ref{e:fprule1}), 
$min(M_1, M_2)$ yields a vector whose elements are the minimal between the corresponding elements in $M_1$ and $M_2$. It essentially represents the intersection of the two masks. 
The hyperparameter $\beta\in (0,1)$
stands for a threshold to distinguish natural and injected triggers. 
Intuitively, the condition asserts that if the size of mask intersection is larger than $\beta$ times the minimum size of the two masks, meaning the two have a large part in common,
the trigger is natural.
If all the reverse engineered triggers for a model are natural, the model is considered clean.



\smallskip
\noindent
{\bf Additional Validation Checks.}
In practice, due to the uncertainty in the stochastic symmetric 
differencing algorithm, the presence of local minimums in optimization, and the small number of available clean samples, 
$M_1$ and $M_2$ may not have a lot in common. However, 
they should nonetheless  satisfy the semantic constraint  that both should denote natural feature differences of the victim and target classes if the trigger is natural. Therefore, we propose an additional cross-validation check that tests if functionally $M_1$ and $M_2$ can take each other's place in satisfying constraints (\ref{e:cns1}) and (\ref{e:cns2}).
In particular, 
while $M_1$ is derived by comparing the victim class and the target class clean samples, we copy the feature maps indicated by $M_1$ between the victim samples and their compromised versions with trigger and check if the intended class flipping can be induced; similarly, while $M_2$ is derived by comparing the victim class samples and their compromised versions, we copy the feature maps indicated by $M_2$ between the victim clean samples and the target clean samples  to see if the intended class flipping can be induced. 
If so, the two are functionally similar and the trigger is natural.
The check is formulated as follows.
\begin{align}\label{e:fprule2}
\begin{split}
&    Acc( h(g(X_V)\cdot M_2 + g(X_T)\cdot \neg M_2), V) > \gamma\  \land\\
&    Acc( h(g(X_T)\cdot M_2 + g(X_V)\cdot \neg M_2), T) > \gamma\  \land\\
&    Acc( h(g(X_V)\cdot M_1 + g(X_V+t)\cdot \neg M_1), V) > \gamma \ \land\\
&    Acc(h(g(X_V+t)\cdot M_1 + g(X_V)\cdot \neg M_1), T) > \gamma \\
\end{split}
\end{align}
Here, $Acc()$ is a function to evaluate prediction accuracy on a set of samples and $\gamma$ a threshold (0.8 in the paper).
We use $g(X_V)$ to denote applying $g$ on each sample in $X_V$ for representation simplicity.

\section{Evaluation}
\label{sec:eval}
\sname{} is implemented on PyTorch~\cite{pytorch}. We will release the code upon publication. We conduct a number of experiments, including evaluating \sname{} on TrojAI rounds 2-4 datasets (with round 4 the latest) and a number of ImageNet pre-trained and trojaned models. 
We also apply \sname{} to detect composite backdoors and reflection backdoors in models for CIFAR10, and hidden-trigger backdoors in models for ImageNet.  
These are semantic backdoors, meaning that their triggers are natural objects/features instead of noise-like patches/watermarks. They may be large and complex. We study the performance of \sname{} in different settings, and compare with 8 baselines that make use of simple L2 distance, attribution/interpretation techniques, and one-sided (instead of symmetric) analysis. At the end, we design an adaptive attack and evaluate \sname{} against it.

\noindent
{\bf Datasets, Models and Hyperparameter Setting.}
Note that the data processed by \sname{} are trained models. 
We use TrojAI rounds 2-4 training and test datasets~\cite{TrojAI:online}. \sname{} {\em does not} require training and hence we use both training and test sets as regular datasets in our experiments. TrajAI round 2 training set has 552 clean models and 552 trojaned models with 22 structures.
It has two types of backdoors: polygons and Instagram filters. 
Round 2 test set has 72 clean and 72 trojaned models. Most performers had difficulties for round 2 due to the prevalence of natural triggers. IARPA hence introduced adversarial training~\cite{madry2017towards,wong2020fast}
in round 3 to enlarge the distance between classes and suppress natural triggers. 
Round 3 training set has 504 clean and 504 trojaned models and the test set has 144 clean and 144 trojaned models. 
In round 4, triggers may be position dependent, meaning that they only cause misclassification when stamped at a specific position inside the foreground object. A model may have multiple backdoors.
The number of clean images provided is reduced from 10-20 (in rounds 2 and 3) to 2-5. 
Its training set has 504 clean and 504 trojaned models and the test set has 144 clean and 144 trojaned models.
Training sets were evaluated on our local server whereas test set evaluation was done remotely by IARPA on their server. At the time of evaluation, the ground truth of test set models was unknown.

We also use a number of models on ImageNet. They have the VGG, ResNet and DenseNet structures. We use 7 trojaned models from~\cite{liu2019abs} and 17 pre-trained clean models from torchvision zoo~\cite{torchvision:online}.
In the composite attack experiment, we use 25 models on CIFAR10. 
They have the Network in Network structure. Five of them are trojaned (by composite backdoors) using the code provided at~\cite{lin2020composite}. We further mix them with 20 pre-trained models from~\cite{liu2019abs}.
In the reflection attack experiment, we use 25 Network in Network models on CIFAR10. Five of them are trojaned (by reflection backdoors) using the code provided at~\cite{liu2020reflection}. The 20 pre-trained clean models are from~\cite{liu2019abs}. 
In the hidden trigger attack experiment, we use 23 models on ImageNet. Six of them are trojaned with hidden-triggers using the code at~\cite{saha2020hidden} and they are mixed with 17 pre-trained clean models from torchvision zoo~\cite{torchvision:online}.

The other settings can be found in Appendix~\ref{sec:setting}.

\subsection{Experiments on TrojAI and ImageNet Models}

\begin{table*}[]
\footnotesize
\centering
\caption{Effectiveness of \sname; (T:276,C:552) means that there are 276 trojaned models and 552 clean models}
\label{t:effectiveness}
\setlength{\tabcolsep}{3pt}
\begin{tabular}{crrrrrrrlrrrrrrrlrrrrrrrrrrr}
\toprule
 & \multicolumn{7}{c}{TrojAI R2} & \multicolumn{1}{c}{} & \multicolumn{7}{c}{TrojAI R3} & \multicolumn{1}{c}{} & \multicolumn{7}{c}{TrojAI R4} & \multicolumn{1}{c}{} & \multicolumn{3}{c}{ImageNet} \\
 \cmidrule{2-8} \cmidrule{10-16} \cmidrule{18-24} \cmidrule{26-28}
 & \multicolumn{3}{c}{\begin{tabular}[c]{@{}c@{}}Polygon trigger\\ (T:276,C:552)\end{tabular}} & \multicolumn{1}{c}{} & \multicolumn{3}{c}{\begin{tabular}[c]{@{}c@{}}Filter trigger\\ (T:276,C:552)\end{tabular}} & \multicolumn{1}{c}{} & \multicolumn{3}{c}{\begin{tabular}[c]{@{}c@{}}Polygon trigger\\ (T:252,C:504)\end{tabular}} & \multicolumn{1}{c}{} & \multicolumn{3}{c}{\begin{tabular}[c]{@{}c@{}}Filter trigger\\ (T:252,C:504)\end{tabular}} & \multicolumn{1}{c}{} & \multicolumn{3}{c}{\begin{tabular}[c]{@{}c@{}}Polygon trigger\\ (T:143,C:504)\end{tabular}} & \multicolumn{1}{c}{} & \multicolumn{3}{c}{\begin{tabular}[c]{@{}c@{}}Filter trigger\\ (T:361,C:504)\end{tabular}} & \multicolumn{1}{c}{} & \multicolumn{3}{c}{\begin{tabular}[c]{@{}c@{}}Patch Trigger\\ (T:7, C:17)\end{tabular}} \\
 \cmidrule{2-4} \cmidrule{6-8} \cmidrule{10-12} \cmidrule{14-16} \cmidrule{18-20} \cmidrule{22-24} \cmidrule{26-28} 
 & \multicolumn{1}{c}{TP} & \multicolumn{1}{c}{FP} & \multicolumn{1}{c}{Acc} & \multicolumn{1}{c}{} & \multicolumn{1}{c}{TP} & \multicolumn{1}{c}{FP} & \multicolumn{1}{c}{Acc} & \multicolumn{1}{c}{} & \multicolumn{1}{c}{TP} & \multicolumn{1}{c}{FP} & \multicolumn{1}{c}{Acc} & \multicolumn{1}{c}{} & \multicolumn{1}{c}{TP} & \multicolumn{1}{c}{FP} & \multicolumn{1}{c}{Acc} & \multicolumn{1}{c}{} & \multicolumn{1}{c}{TP} & \multicolumn{1}{c}{FP} & \multicolumn{1}{c}{Acc} & \multicolumn{1}{c}{} & \multicolumn{1}{c}{TP} & \multicolumn{1}{c}{FP} & \multicolumn{1}{c}{Acc} & \multicolumn{1}{c}{} & \multicolumn{1}{c}{TP} & \multicolumn{1}{c}{FP} & \multicolumn{1}{c}{Acc} \\
Vanilla ABS & 254 & 218 & 0.710 &  & 260 & 293 & 0.626 &  & 235 & 208 & 0.702 &  & 213 & 334 & 0.528 &  & 137 & 355 & 0.442 &  & 331 & 376 & 0.531 &  & 7 & 7 & 0.708 \\
Inner L2 & 188 & 93 & 0.782 &  & 153 & 123 & 0.703 &  & 210 & 111 & 0.798 &  & 133 & 123 & 0.680 &  & 73 & 137 & 0.680 &  & 208 & 217 & 0.572 &  & 7 & 0 & 1 \\
Inner RF & 192 & 76 & 0.807 &  & 196 & 101 & 0.781 &  & 159 & 46 & 0.816 &  & 153 & 110 & 0.724 &  & 133 & 265 & 0.575 &  & 330 & 353 & 0.556 &  & 7 & 0 & 1 \\
IG & 172 & 29 & 0.840 &  & 192 & 66 & 0.818 &  & 162 & 58 & 0.804 &  & 52 & 41 & 0.681 &  & 84 & 53 & 0.827 &  & 210 & 87 & 0.725 &  & 5 & 0 & 0.917 \\
Deeplift & 152 & 11 & 0.837 &  & 189 & 21 & 0.869 &  & 162 & 59 & 0.803 &  & 78 & 67 & 0.681 &  & 84 & 54 & 0.825 &  & 203 & 54 & 0.755 &  & 6 & 0 & 0.958 \\
Occulation & 173 & 24 & 0.847 &  & 207 & 47 & 0.860 &  & 164 & 58 & 0.807 &  & 78 & 66 & 0.683 &  & 85 & 52 & 0.830 &  & 251 & 107 & 0.749 &  & 7 & 3 & 0.875 \\
NE & 180 & 58 & 0.814 &  & - & - & - &  & 187 & 72 & 0.819 &  & - & - & - &  & 59 & 72 & 0.759 &  & - & - & - &  & 7 & 4 & 0.833 \\
1-sided(V to T) & 157 & 19 & 0.833 &  & 195 & 33 & 0.862 &  & 202 & 62 & 0.852 &  & 153 & 51 & 0.802 &  & 107 & 82 & 0.818 &  & 236 & 50 & 0.798 &  & 7 & 0 & 1 \\
1-sided(T to V) & 134 & 4 & 0.824 &  & 158 & 18 & 0.835 &  & 187 & 50 & 0.848 &  & 134 & 27 & 0.808 &  & 102 & 56 & 0.850 &  & 179 & 9 & 0.779 &  & 1 & 1 & 0.958 \\
\sname & 198 & 19 & 0.883 &  & 204 & 32 & 0.874 &  & 200 & 46 & 0.870 &  & 149 & 39 & 0.812 &  & 105 & 53 & 0.859 &  & 242 & 46 & 0.809 &  & 7 & 0 & 1 \\
\bottomrule
\end{tabular}
\end{table*}

In the first experiment, we evaluate \sname{} on TrojAI rounds 2-4 training sets and the ImageNet models. We do not include TrojAI 
test sets in this experiment as the test sets are hosted on an IARPA server and do not provide ground-truth information. One can only submit a solution to the server, which returns the overall scanning accuracy and cross-entropy loss.
Here we use ABS as the upstream scanner as it is much faster than NC. 

\begin{figure}
    \centering                                                   
    \footnotesize 

     \includegraphics[width=0.3\textwidth]{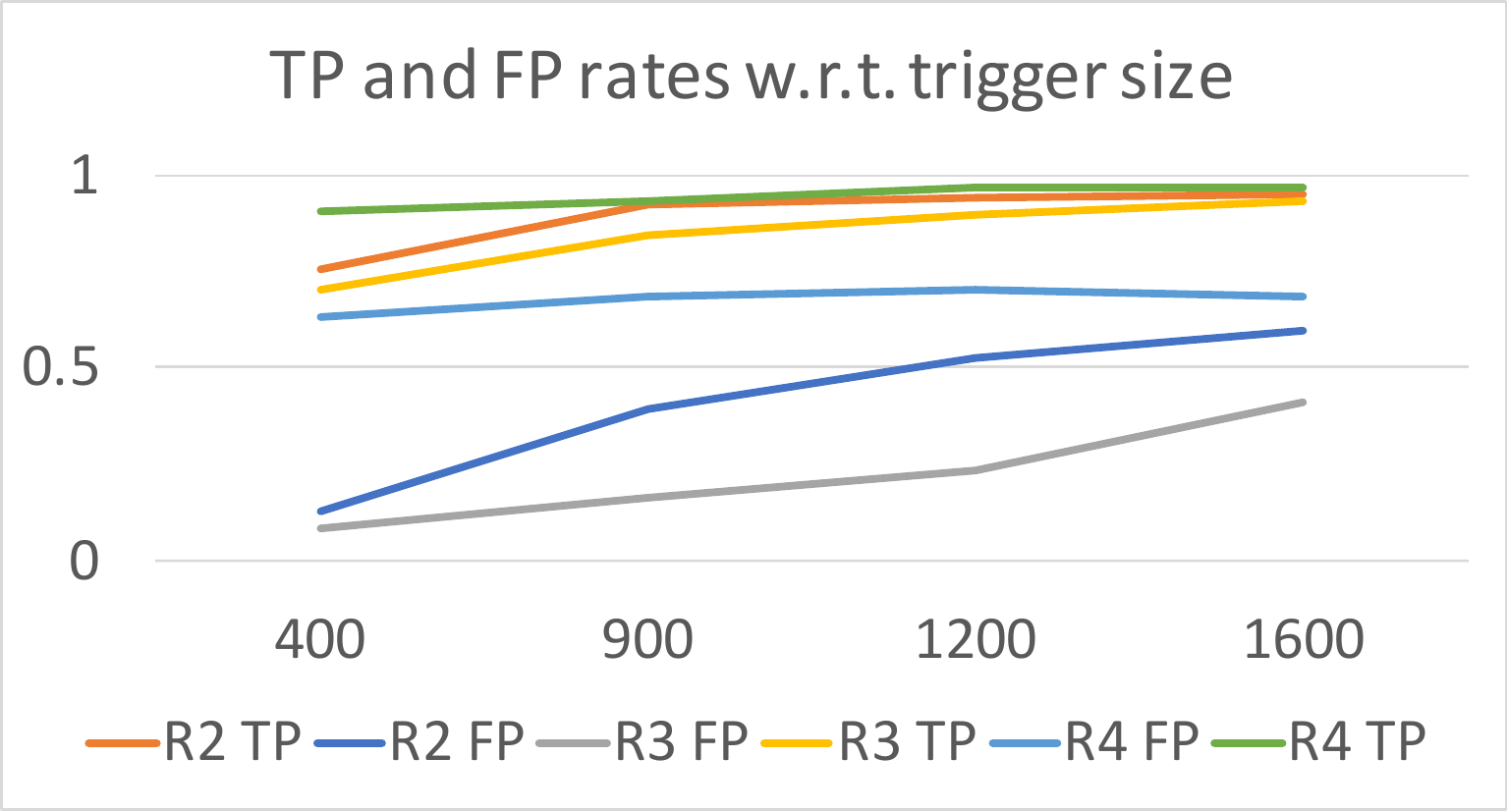}

    \caption{
    Rounds 2-4 true positive rates (TPs) and false positive rates (FPs) versus trigger size (in pixels) by ABS 
    } 

   \label{f:fp_tp_size}
\end{figure}

A critical setup for scanners that produce triggers, such as ABS and NC, is the maximum trigger size. A large value enables detecting injected backdoors with large triggers, while producing a lot of natural triggers and hence false positives. Fig.~\ref{f:fp_tp_size} studies how the true positives (TPs) and false positives (FPs)  change with different trigger bounds in ABS, on the TrojAI rounds 2-4 training sets. Observe that both grow with the trigger size. 
Observe that there is a lower FP rate in round 3 (compared to round 2), illustrating the effect of adversarial training, although the number is still large when the trigger size is large. 
Round 4 has the highest FP rate because the number of clean images available is decreased and it is hence very easy for scanners to find (bogus) triggers that can induce misclassification on all the available images. 

Based on the study, we use the trigger size bound 900 pixels for round 2, 1600 pixels for round 3, and 1200 pixels for round 4 for our experiment 
such that the upstream scanner does not miss many true positives to begin with and we can stress test \sname.

\noindent
\textbf{Baselines.} In the experiment, we compare \sname{} against 8 baselines. 
The first baseline is using L2 distance of inner activation between $V+t$ and $T$.
Such a distance for a natural trigger is supposed to be smaller than that of an injected trigger (more in Section~\ref{s:motivation}).
We use unsupervised learning to report the best separation.
In the second baseline, we use half of the models to train a random forest classifier based on the inner activations and logits values to distinguish natural and injected triggers and test it on the other half.
Specifically, the classifier takes the L2 distance between $V+t$ and $T$, L2 distance between $T+t$ and $T$, L2 distance between $V$ and $T$, logits of $V$, logits of $V+t$ and logits of $T$.
The third baseline uses {\em integrated gradients (IG)}~\cite{sundararajan2017axiomatic}, an attribution technique, to find important neurons for $V+t$ and for $T$ and then apply the aforementioned L2 distance comparison on the 10\% most important neurons (more in Section~\ref{s:motivation}). 
Originally, integrated gradients were used in model explanation and identify important pixels. We adapt it to work on inner layers and identify important neurons. 
The next three baselines are similar to the third except having different methods to identify important neurons.
Specifically, the fourth baseline uses {\em Deeplift}~\cite{shrikumar2016not}, the fifth uses {\em Occlusion}~\cite{ancona2018towards} and the sixth uses {\em Network Dissection (NE)}~\cite{bau2017network}. 
For baselines 4-7, we use unsupervised learning to find the best separation (of natural and injected backdoors).
We will release the settings together with our system upon publication.
\sname{} is symmetric. To study the necessity of symmetry, the seventh and eighth baselines are one-sided versions of \sname, that is, requiring satisfying either constraint (\ref{e:cns1}) or (\ref{e:cns2}) in Section~\ref{sec:design}.

The results are shown in Table~\ref{t:effectiveness}. 
The first column shows the methods.
The first method is the vanilla ABS.
Columns 2-4 show the results for TrojAI round 2 models with polygon backdoors. Column 2 shows the number of true positives (TPs). Note that there are 276 trojaned models with polygon backdoors. As such the vanilla ABS having 254 TPs means it has 22 false negatives.
Column 3 shows the number of false positives (FPs) out of the 552 clean models. Column 4 shows the overall detection accuracy (on the total 552+276=828 models).
Columns 5-7 show the results for round 2 models with Instagram filter backdoors. ABS uses a different method for filter backdoors. Instead of reverse engineering a pixel patch, it reverse engineers a one-layer kernel denoting general filter transformation~\cite{liu2019abs}. Hence, we separate the evaluation of \sname{} on the two kinds of backdoors. Note that the accuracy is computed considering the same 552 clean models. The overall results (for all kinds of backdoors) on the leaderboard are presented later.
Columns 8-13 show the results for round 3 and columns 14-19 for round 4.  
Columns 14-16 show the results for ImageNet patch attack.

\begin{table*}[]
\caption{TrojAI leaderboard results }
\label{t:trojai}
\footnotesize
\centering
\begin{tabular}{crrrrrrrr}
\toprule
 & \multicolumn{2}{c}{Round 2} & \multicolumn{1}{c}{} & \multicolumn{2}{c}{Round 3} & \multicolumn{1}{c}{} & \multicolumn{2}{c}{Round 4} \\
 \cmidrule{2-3} \cmidrule{5-6} \cmidrule{8-9}
 & \multicolumn{1}{c}{CE loss} & \multicolumn{1}{c}{ROC-AUC} & \multicolumn{1}{c}{} & \multicolumn{1}{c}{CE loss} & \multicolumn{1}{c}{ROC-AUC} & \multicolumn{1}{c}{} & \multicolumn{1}{c}{CE loss} & \multicolumn{1}{c}{ROC-AUC} \\
ABS only & 0.685 & 0.736 &  & 0.541 & 0.822 &  & 0.894 & 0.549 \\
ABS+\sname{} & 0.324 & 0.892 &  & 0.323 & 0.9 &  & 0.322 & 0.902 \\
Deficit from top & 0 & 0 &  & 0.023 & -0.012 &  & 0 & 0 \\
\bottomrule
\end{tabular}
\vspace{-0.2in}
\end{table*}

The results show that the vanilla ABS has a lot of FPs (in order not to lose TPs) and \sname{} 
can substantially reduce the FPs 
by 78-100\%
with the cost of increased FNs (i.e., losing TPs) by 0-30\%. The overall detection accuracy improvement (from vanilla ABS) is 17-41\% 
across the datasets.
Also observe that \sname{} consistently outperforms all the baselines, especially the non-\sname{} ones.
Attribution techniques can remove a lot of natural triggers indicated by the decrease of FPs. However, they preclude many injected triggers (TPs) as well, leading to inferior performance. The missing entries for NE are because it requires an input region to decide important neurons, rendering it inapplicable to filters that are pervasive.
Symmetric \sname{} outperforms the one-sided versions, suggesting the need of symmetry.

\noindent
{\bf Results on TrojAI Leaderboard (Test Sets).}
\label{sec:leaderboard}
In the first experiment, we evaluate \sname{} in a most challenging setting, which causes the upstream scanner to produce the largest number of natural triggers (and also the largest number of true injected triggers). 
TrojAI allows performers to tune hyperparameters based on the training sets. We hence fine-tune our ABS+\sname{} pipeline, searching for the best hyperparameter settings such as maximum trigger size, optimization epochs, $\alpha$, $\beta$, and $\gamma$, on the training sets and then evaluate the tuned version on the test sets.  
Table~\ref{t:trojai} shows the results.
In two of the three rounds, our solution achieved the top performance\footnote{TrojAI ranks solutions based on the cross-entropy loss of scanning results. Intuitively, the loss increases when the model classification diverges from the ground truth. Smaller loss suggests better performance~\cite{TrojAI:online}. Past leaderboard results can be found at~\cite{pastleaderboards}. }.
In round 3, it ranked number 2 and the results are comparable to the top performer. In addition, it beat the IARPA round goal (i.e., cross-entry loss lower than 
0.3465
) for all the three rounds. 
We also want to point out that \sname{} is not a stand-alone defense technique. Hence, we do not directly compare with existing end-to-end defense solutions. Our performance on the leaderboard, especially for round 2 that has a large number of natural backdoors and hence caused substantial difficulties for most performers\footnote{Most performers had lower than 0.80 accuracy in round 2.}, suggests the contributions of \sname. As far as we know, many existing solutions such as~\cite{liu2019abs, wang2019neural, kolouri2020universal, tang2019demon, suciu2018does, erichson2020noise, jha2019attribution, chen2019deepinspect, sikka2020detecting} have been tested in the competition by different performers. 

\noindent
{\bf Runtime. }
On average, \sname{} takes 12s to process a trigger, 95s to process a model. ABS takes 337s to process a model, producing 8.5 triggers on average.


\noindent
{\bf Effects of Hyperparameters.}
We study \sname{} performance with various settings, including the different layer to which  \sname{} is applied, different trigger size settings (in the upstream scanner) and different SSIM score bound (in the upstream filter backdoor scanning to ensure the
generated kernel does not over-transform an input), and the 
$\alpha$, $\beta$, and $\gamma$ settings of  \sname{}. The results are in Appendix~\ref{sec:hyperparameter}. They show that  \sname{} is reasonably stable with various settings.

\begin{table}[]
\caption{\sname{} with different upstream scanners}
\label{t:upstream}
\centering
\footnotesize
\setlength{\tabcolsep}{2    pt}
\begin{tabular}{crrrrrrrrrrrr}
\toprule
 & \multicolumn{5}{c}{Vanilla} & \multicolumn{1}{c}{} & \multicolumn{6}{c}{+\sname} \\
 \cmidrule{2-6} \cmidrule{8-13}
 & \multicolumn{1}{c}{TP} & \multicolumn{1}{c}{T} & \multicolumn{1}{c}{FP} & \multicolumn{1}{c}{C} & \multicolumn{1}{c}{Acc} & \multicolumn{1}{c}{} & \multicolumn{1}{c}{TP} & \multicolumn{1}{c}{T} & \multicolumn{1}{c}{FP} & \multicolumn{1}{c}{C} & \multicolumn{1}{c}{Acc} & \multicolumn{1}{c}{Acc Inc} \\
NC & 180 & 252 & 332 & 552 & 0.483 &  & 127 & 252 & 73 & 552 & 0.732 & 0.249 \\
SRI-RE & 164 & 252 & 272 & 552 & 0.536 &  & 112 & 252 & 97 & 552 & 0.685 & 0.149 \\
SRI-CLS & 120 & 146 & 17 & 158 & 0.858 &  & 119 & 146 & 9 & 158 & 0.882 & 0.024 \\
\bottomrule
\end{tabular}
\end{table}

\subsection{Using \sname{} with Different Upstream Scanners}

In this experiment, we use \sname{} with different upstream scanners, including Neural Cleanse (NC)~\cite{wang2019neural} and 
the Bottom-pp-Top-down method
by the SRI team in the TrojAI competition~\cite{sriscanner}.  The latter
has two sub-components, trigger generation and a classifier that makes use of 
features collected from the trigger generation process.
We created two scanners out of their solution. In the first one, 
we apply \sname{} on top of their final classification results (i.e., using \sname{} as a refinement). We call it SRI-CLS. In the second one, we
apply \sname{} right after their trigger
generation. We have to replace their classifier with the simpler unsupervised learning (i.e., finding the best separation) as adding \sname{} changes the features and nullifies their original classifier. We call it SRI-RE. We use the round 2 clean models and models with polygon triggers to conduct the study as NC does not handle Instagram filter triggers. For SRI-CLS, 
the training was on 800 randomly selected models and testing was on the remaining 146 trojaned models and 158 clean models. The other scanners do not require training. 
The results are shown in Table~\ref{t:upstream}.
The T and C columns stand for the number of trojaned and clean models used in testing, respectively.
 Observe that the vanilla NC identifies 180 TPs and 332 FPs with the accuracy of 44.7\%. With \sname,  the FPs are reduced to 73 (81.1\% reduction) and the TPs become 127 (29.4\% degradation). 
The overall accuracy improves from 44.7\% to 70.8\%.
The improvement for SRI-RE is from 53.6\% to 68.5\%. The improvement for SRI-CLS is relative less significant. That is because 0.882 accuracy is already very close to the best performance for this round.
The results show that \sname{} can consistently improve upstream scanner performance.
Note that the value of \sname{} lies in suppressing false warnings. It offers little help if the upstream scanner has substantial false negatives.
In this case, users may want to tune the upstream scanner to have minimal false negatives and then rely on the downstream \sname{} to prune the resulted false positives like we did in the ABS+\sname{} pipeline.


\subsection{Study of Failing Cases by \sname{}}
\label{sec:case}
\begin{figure}
    \centering                                                   
    \footnotesize
    \subfigure[Victim]{
        \includegraphics[width=0.1\textwidth]{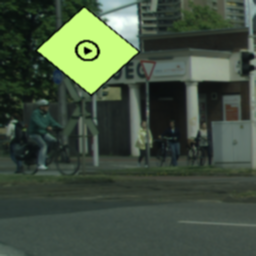}
    }                              
    ~
    \subfigure[Target]{
        \includegraphics[width=0.1\textwidth]{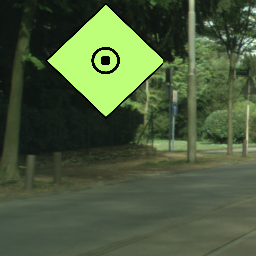}
    }  
    ~
    \subfigure[Trigger]{
        \includegraphics[width=0.1\textwidth]{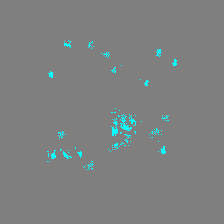}
    }
    \caption{\sname{} fails to prune a natural backdoor} 

   \label{f:naturalexample}
\end{figure}

\begin{figure}
    \centering                                                   
    \footnotesize
    \subfigure[Victim]{
        \includegraphics[width=0.1\textwidth]{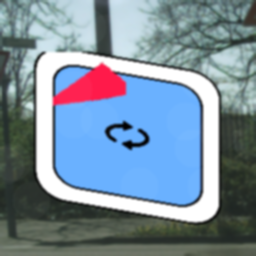}
    }                              
    ~
    \subfigure[Target]{
        \includegraphics[width=0.1\textwidth]{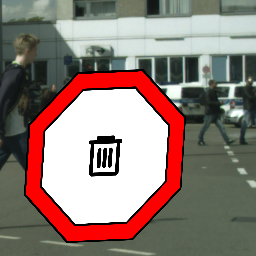}
    }  
    ~
    \subfigure[Trigger]{
        \includegraphics[width=0.1\textwidth]{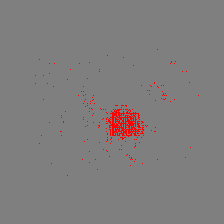}
    }
    \caption{\sname{} misclassifies a trojaned model to clean} 

   \label{f:injectexample}
\end{figure}

\noindent
According to Table~\ref{t:effectiveness}, \sname{} cannot prune all the natural triggers (causing FPs) and it may mistakenly prune injected triggers (causing FNs).  
Here, we study two cases: one demonstrating why \sname{} fails to detect a natural trigger and the other demonstrating why \sname{} misclassifies an injected trigger to natural. 
Fig.~\ref{f:naturalexample} shows a clean model with a natural trigger but \sname{} fails to prune it, with (a) and (b) the victim and target classes, respectively, and (c) the natural trigger by ABS. 
Observe that the victim and target classes are really close. Even the central symbols look similar. As such, small and arbitrary input perturbations as those in (c) may be sufficient to induce misclassification.  Such perturbations may not resemble any of the distinguishing features between the two classes at all, rendering \sname{} ineffective. 
Fig.~\ref{f:injectexample} shows a trojaned model that \sname{} considers clean, with (a) and (b) the victim class with the injected trigger (ground truth provided by IARPA) and the target class, respectively, and (c) the trigger generated by ABS. 
Observe that a strong distinguishing feature of the victim and target classes is the red versus the white borders. The injected trigger happens to be a red polygon, which shares a lot of commonality with the differential features of the classes, rendering \sname{} ineffective. 


\subsection{Detecting Semantic Backdoor Attacks}
\label{sec:composite}

While early backdoor attacks on image classifiers used noise-like pixel patches/watermarks that are small, recent backdoor attacks showed that models can be trojaned with natural objects and features that may be large and complex. As such, scanners that generate small triggers to determine if a model has backdoor become ineffective. A possible solution is to enlarge the size bound such that the injected large triggers can be generated. However, this entails a lot of false positives as it admits many natural triggers. In this experiment, we show that by enlarging the trigger bound of ABS and using \sname{} to prune false positives, we can detect composite backdoors~\cite{lin2020composite}, hidden-trigger backdoors~\cite{saha2020hidden}, and reflection backdoors~\cite{liu2020reflection}.

\smallskip
\noindent
{\bf Detecting Composite Backdoor.}
Composite backdoor uses composition of existing benign features as triggers (see Fig.~\ref{f:compositeexamples} in Section~\ref{sec:intro}). 
The experiment is on 5 models trojaned by the attack and 20 clean models on CIFAR10. We set the trigger size to 600 pixels in order to reverse engineer the large trigger features used in the attack. In Table~\ref{t:composite}, our results show that we can achieve 0.84 accuracy (improved from 0.2 by vanilla ABS), reducing the false positives from 20 to 4. It shows the potential of \sname{}.

\begin{figure}
    \centering                                                   
    \footnotesize
    \subfigure[Natural]{
    \begin{minipage}[c]{0.5in}
       \center
        \includegraphics[width=0.5in]{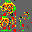}
    \end{minipage}
    }
    ~
    \subfigure[T: Dog]{
    \begin{minipage}[c]{0.5in}
       \center
        \includegraphics[width=0.5in]{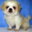}
    \end{minipage}
    }
    ~
    \subfigure[Composite]{
    \begin{minipage}[c]{0.6in}
       \center
        \includegraphics[width=0.5in]{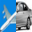}
    \end{minipage}
    }
    ~
    \subfigure[Injected]{
    \begin{minipage}[c]{0.5in}
       \center
        \includegraphics[width=0.5in]{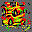}
    \end{minipage}
    }
    ~
    \subfigure[T: Bird]{
    \begin{minipage}[c]{0.5in}
       \center
        \includegraphics[width=0.5in]{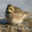}
    \end{minipage}
    }   
                             
    \caption{Example of natural and injected composite trigger
    } 

   \label{f:compositetrigger}
\end{figure}

Fig.~\ref{f:compositetrigger} shows a natural trigger and an injected composite trigger. 
Figures (a) and (b) show a natural trigger (generated by ABS) with the target label dog and a sample from the dog class (in CIAFR10), respectively. Observe that the trigger has a lot of dog features (and hence pruned by \sname{}).
Figures (c) and (d) show a composite trigger used during poisoning, which is a combination of car and airplane, and the trigger generated by ABS, respectively.
Figure (e) shows the target label bird. 
 Observe that the reverse engineered trigger has car features (e.g., wheels). \sname{} recognizes it as an injected trigger since it shares very few features with the bird class.

\begin{table}[]
\caption{\sname{} on composite attack, with 20 clean models and 5 trojaned models}
\label{t:composite}
\centering
\footnotesize
\begin{tabular}{crrrrr}
\toprule
 & \multicolumn{1}{c}{TP} & \multicolumn{1}{c}{FP} & \multicolumn{1}{c}{FN} & \multicolumn{1}{c}{TN} & \multicolumn{1}{c}{Acc} \\
 \midrule
 ABS& 5 & 20 & 0 & 0 & 0.2 \\
ABS+\sname{} & 5 & 4 & 0 & 16 & 0.84 \\
\bottomrule
\end{tabular}
\end{table}

\begin{figure}
    \centering                                                   
    \footnotesize
    \subfigure[Trigger]{
    \begin{minipage}[c]{0.6in}
       \center
        \includegraphics[width=0.5in]{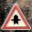}
    \end{minipage}
    }
    ~
    \subfigure[Reflection]{
    \begin{minipage}[c]{0.6in}
       \center
        \includegraphics[width=0.5in]{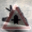}
    \end{minipage}
    }
    ~
    \\
    \subfigure[Natural]{
    \begin{minipage}[c]{0.5in}
       \center
        \includegraphics[width=0.5in]{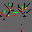}
    \end{minipage}
    }
    ~
    \subfigure[T: Deer]{
    \begin{minipage}[c]{0.5in}
       \center
        \includegraphics[width=0.5in]{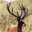}
    \end{minipage}
    }
    ~
    \subfigure[Injected]{
    \begin{minipage}[c]{0.5in}
       \center
        \includegraphics[width=0.5in]{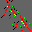}
    \end{minipage}
    }
    ~
    \subfigure[T: Plane]{
    \begin{minipage}[c]{0.5in}
       \center
        \includegraphics[width=0.5in]{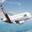}
    \end{minipage}
    }   
                             
    \caption{Example of natural and injected reflection trigger
    } 

   \label{f:reflection}
\end{figure}

\begin{table}
\caption{\sname{} on reflection attack, with 20 clean models and 5 trojaned models}
\label{t:reflection}
\centering
\footnotesize
\begin{tabular}{crrrrr}
\toprule
 & \multicolumn{1}{c}{TP} & \multicolumn{1}{c}{FP} & \multicolumn{1}{c}{FN} & \multicolumn{1}{c}{TN} & \multicolumn{1}{c}{Acc} \\
 \midrule
ABS  & 5 & 17 & 0 & 3 & 0.32 \\
ABS+\sname{} & 5 & 3 & 0 & 17 & 0.88 \\
\bottomrule
\end{tabular}
\vspace{-.2in}
\end{table}

\smallskip
\noindent
{\bf Detecting Reflection Backdoor.}
Reflection may occur when taking picture behind a glass window. Reflection backdoors uses the reflection of an image as the trojan trigger. Figure~\ref{f:reflection} (a) and (b) shows an image of triangle sign and its reflection on an image of airplane.  Reflection attack uses the reflection of a whole image as the trigger, which is large and complex. 
We evaluate ABS+\sname{} on 5 models trojaned with reflection backdoors and 20 clean models on CIFAR10. The trigger size bound is set to 256 (very large for CIFAR images). Table~\ref{t:reflection} shows that we can achieve 0.88 accuracy (improved from 0.32 by ABS), reducing the false positives from 17 to 3. 

Figure~\ref{f:reflection} (c) and (d) show a natural trigger (generated by ABS) with the target label deer and a sample from the deer class. Observe that the trigger resembles deer antlers.
Figure~\ref{f:reflection} (e) and (f) show the trigger generated by ABS with the target class airplane. Observe that the generated trigger has (triangle) features of the real trigger  shown in Figure~\ref{f:reflection} (a) and (b). \sname{} classifies it as injected as it shares few features with airplane (Figure~\ref{f:reflection} (f)).

\begin{figure}
    \centering                                                   
    \footnotesize
    \subfigure[Trigger]{
    \begin{minipage}[c]{0.7in}
       \center
        \includegraphics[width=0.7in]{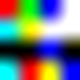}
    \end{minipage}
    }
    ~
    \subfigure[Stamped]{
    \begin{minipage}[c]{0.7in}
       \center
        \includegraphics[width=0.7in]{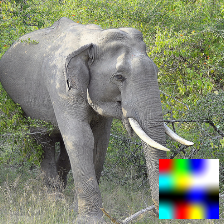}
    \end{minipage}
    }
    \\
    ~
    \subfigure[Natural]{
    \begin{minipage}[c]{0.7in}
       \center
        \includegraphics[width=0.7in]{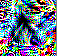}
    \end{minipage}
    }
    ~
    \subfigure[T: Jeans]{
    \begin{minipage}[c]{0.7in}
       \center
        \includegraphics[width=0.7in]{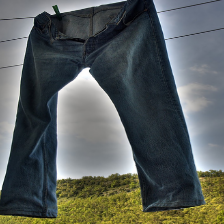}
    \end{minipage}
    }
    ~
    \subfigure[Injected]{
    \begin{minipage}[c]{0.7in}
       \center
        \includegraphics[width=0.7in]{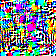}
    \end{minipage}
    }
    ~
    \subfigure[T: terrier]{
    \begin{minipage}[c]{0.7in}
       \center
        \includegraphics[width=0.7in]{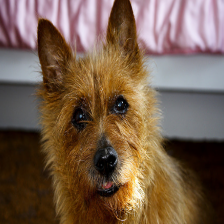}
    \end{minipage}
    }   
                             
    \caption{Example of natural and injected hidden-triggers. The triggers have been enlarged for readability.
    } 

   \label{f:hidden}
\vspace{-.1in}
\end{figure}

\smallskip
\noindent
{\bf Detecting Hidden-trigger Backdoor.}
Hidden-trigger attack does not directly use trigger to poison training data. Instead, it introduces perturbation on the images of target label such that the perturbed images induce similar inner layer activations to the images of victim label stamped with the trigger. Since the inner layer activations represent features, the model picks up the correlations between trigger features  and the target label. Thus images stamped with the trigger are misclassified to the target label at 
test time.  The attack is a clean label attack.
Since the trigger is not explicit, the attack is more stealthy compared to data poisoning.
On the other hand, the trojaning process is more difficult, demanding larger triggers, causing problems for existing scanners.
For example, it requires the trigger size to be 60$\times$60 for ImageNet to achieve a high attack success rate. Figure~\ref{f:hidden} (a) and (b) show a trigger and an ImageNet sample stamped with the trigger, with the target label terrier dog shown in (f). 

We evaluate ABS+\sname{} on 6 models trojaned with hidden-triggers and 17 clean models on ImageNet. The trigger size bound for ABS is set to 4000.
Table~\ref{t:hidden} shows that we can achieve 0.82 accuracy (improved from 0.3 by ABS), reducing the number of false positives from 16 to 3.
Figure~\ref{f:hidden} (c) and (d) show a natural trigger generated by ABS and its target label Jeans. Observe that the center part of natural trigger resembles a pair of jeans pants. Since the natural trigger mainly contains features of the target label, it is pruned by \sname.
Figure~\ref{f:hidden} (e) shows the generated trigger by ABS for the target label (f). The trigger has few in common with the target label and thus is classified as injected by \sname{}. 

\begin{table}[t]
\caption{\sname{} on hidden-trigger attack, with 17 clean models and 6 trojaned models}
\label{t:hidden}
\centering
\footnotesize
\begin{tabular}{crrrrr}
\toprule
 & \multicolumn{1}{c}{TP} & \multicolumn{1}{c}{FP} & \multicolumn{1}{c}{FN} & \multicolumn{1}{c}{TN} & \multicolumn{1}{c}{Acc} \\
 \midrule
ABS  & 6 & 16 & 0 & 1 & 0.30 \\
ABS+\sname{} & 5 & 3 & 1 & 14 & 0.82 \\
\bottomrule
\end{tabular}
\vspace{-.1in}
\end{table}

\begin{table}[]
\vspace{-.1in}
\caption{Adaptive Attack}
\label{t:adaptive}
\footnotesize
\tabcolsep=5pt
\centering
\begin{tabular}{crrrrr}
\toprule
Weight of adaptive loss & \multicolumn{1}{c}{1} & \multicolumn{1}{c}{10} & \multicolumn{1}{c}{100} & \multicolumn{1}{c}{1000} & \multicolumn{1}{c}{10000} \\
\midrule
Acc & 0.89 & 0.88 & 0.87 & 0.82 & 0.1 \\
Asr & 0.99 & 0.99 & 0.99 & 0.97 & - \\
FP/ \# of clean models & 0 & 0.2 & 0.2 & 0.65 & - \\
TP/ \# of trojaned models & 1 & 1 & 1 & 1 & - \\
\bottomrule
\end{tabular}
\end{table}


\subsection{Adaptive Attack}
\label{sec:adaptive}

\sname{} is part of a defense technique and hence vulnerable to adaptive attack.
We devise an adaptive attack that forces the internal activations of victim class inputs embedding the trigger to resemble the activations of the clean target class inputs such that \sname{} cannot distinguish the two.
In particular, we train a Network in Network model on CIFAR10 with a given 8$\times$8 patch as the trigger. 
In order to force the inner activations of images stamped with the trigger to resemble those of target class images, 
we design an adaptive loss which is to minimize the differences between the two.
In particular, we measure the differences of the means and standard deviations of feature maps.
During training, we add the adaptive loss to the normal cross-entropy loss. The effect of adaptive loss is controlled by a weight value, which essentially controls the strength of attack as well.
Besides the adaptively trojaned model, we also train 20 clean models on CIFAR10 to see if ABS+\sname{} can distinguish the trojaned and clean models.

The results are shown in Table~\ref{t:adaptive}. The first row shows the adaptive loss weight. A larger weight value indicates stronger attack.
The second row shows the trojaned model's accuracy on clean images. The third row shows the attack success rate of the trojaned model. The fourth row shows the FP rate.
The fifth row shows the TP rate. 
Observe while ABS+\sname{} does not miss trojaned models, its FP rate grows with the strength of attack. When the weight value is 1000, the FP rate becomes 0.65, meaning \sname{}  is no longer effective. However, the model accuracy also degrades a lot in this case.

\subsection{Fixing Models with Injected and Natural Backdoors}
\label{sec:repair_short}
As mentioned in Section~\ref{sec:intro}, 
an important difference between injected and natural backdoors is that the latter is inevitable and difficult to fix.
To make the comparison,
we try to fix 5 benign models and 5 trojaned models on CIFAR10. The trojaned models are trojaned by label-specific data poisoning. 
Here we use unlearning~\cite{wang2019neural} which stamps triggers generated by scanning methods on images of victim label to finetune the model and forces the model to unlearn the correlations between the triggers and the target label. The process is iterative, bounded by the level of model accuracy degradation. 
The level of repair achieved is measured
by the trigger sizes of the fixed model. 
Larger triggers indicate the corresponding backdoors become more difficult to exploit. The trigger size increase rate suggests the difficulty level of repair. 

Table~\ref{t:unlearning} shows the average accuracy and average reverse engineered trigger size before and after fixing the models. All models have
the same repair budget.
We can see that natural triggers have a larger accuracy decrease. Natural trigger size only increases by 34.4 whereas injected trigger size increases by 78, supporting our hypothesis. 
More detailed results can be found in Appendix~\ref{sec:repair}. Note that model repair is not the focus of the paper and trigger size may not be a good metric to evaluate repair success for the more complex  semantic backdoors. The experiment is to provide initial insights. A thorough model repair solution belongs to our future work.

\begin{table}[htbp]
\footnotesize
\centering
\caption{Average trigger size change before and after unlearning}
\label{t:unlearning}
\begin{tabular}{cccccc}
\toprule
 & \multicolumn{2}{c}{Natural Trigger} &  & \multicolumn{2}{c}{Injected Trigger} \\
  \cmidrule{2-3} \cmidrule{5-6}
 & Before & After &  & Before & After \\
Avg Acc & \multicolumn{1}{r}{88.7\%} & \multicolumn{1}{r}{85.9\%} & \multicolumn{1}{r}{} & \multicolumn{1}{r}{86.4\%} & \multicolumn{1}{r}{85.4\%} \\
Avg Trigger Size & \multicolumn{1}{r}{25.8} & \multicolumn{1}{r}{60.2} & \multicolumn{1}{r}{} & \multicolumn{1}{r}{19} & \multicolumn{1}{r}{97} \\ \bottomrule
\end{tabular}
\end{table}

\section{Related Work}
\label{sec:related_work}
\noindent
\textbf{Backdoor Attack.}
Data poisoning~\cite{GuLDG19, chen2017targeted} injects backdoor by changing the label of inputs with trigger.
Clean label attack~\cite{shafahi2018poison, zhu2019transferable, turner2019label, zhao2020clean, saha2020hidden} injects backdoor without changing the data label.
Bit flipping attack~\cite{rakin2020tbt,rakin2019bit} proposes to trojan models by flipping  weight value bits.
Dynamic backdoor~\cite{salem2020dynamic, nguyen2020input} focuses on crafting different triggers for different inputs and breaks the defense's assumption that trigger is universal.
Ren et al.~\cite{pang2020tale} proposed to combine adversarial example generation and model poisoning to improve the effectiveness of both attacks. 
There are also attacks on NLP tasks~\cite{zhang2020trojaning, chen2020badnl, kurita2020weight}, Graph Neural Network~\cite{zhang2020backdoor, xi2020graph}, transfer learning~\cite{rezaei2019target, wang2018great, yao2019latent}, and federated learning~\cite{xie2019dba, wang2020attack, tolpegin2020data,bagdasaryan2020backdoor,fang2020local}.
\sname{} is a general primitive that may be of use in defending these attacks. 

\noindent
\textbf{Backdoor Attack Defense.} 
ULP~\cite{kolouri2020universal} trains a few universal input patterns and a classifier from thousands of benign and trojaned models. 
The classifier predicts if a model has backdoor based on activations of the patterns. 
Xu et al.~\cite{xu2019detecting} proposed to detect backdoor using a meta classifier trained on a set of trojaned and benign models. 
Qiao et al.~\cite{qiao2019defending} proposed to 
reverse engineer the distribution of triggers.
Hunag et al.~\cite{huang2020one} found that trojaned  and clean models react differently to input perturbations.
Cassandra~\cite{zhang2020cassandra} and TND~\cite{wang2020practical} found that universal adversarial examples behave differently on trojaned and clean models and used this observation to detect backdoor.
TABOR~\cite{guo2019tabor} and NeuronInspect~\cite{huang2019neuroninspect} used an AI explanation technique to detect backdoor.
NNoculation~\cite{veldanda2020nnoculation} used broad spectrum random perturbations and GAN based techniques to reverse engineer trigger.
Besides backdoor detection, there are techniques aiming at removing backdoor.
Fine-prune~\cite{liu2018fine} removed backdoors by pruning out compromised neurons. 
Borgnia et al.~\cite{borgnia2020strong} and Zeng et al.~\cite{zeng2020deepsweep} proposed to use data augmentation technique to mitigate backdoor effect. 
Wang et al.~\cite{wang2020certifying} proposed to use randomized smoothing to certify robustness against backdoor attacks.
There are techniques that defend backdoor attacks by data sanitization where they prune out poisoned training inputs~\cite{cao2018efficient, jagielski2018manipulating, mozaffari2014systematic, paudice2018label}. 
There are also techniques that detect if a given input is stamped with trigger~\cite{ma2019nic, tang2019demon, gao2019strip, chen2018detecting, li2020rethinking, liu2017neural, chou2020sentinet,tran2018spectral, fu2020detecting, chan2019poison, du2019robust, veldanda2020nnoculation}. They target a different problem as they require inputs with embedded triggers.
\sname{} is orthogonal to most of these techniques and can serve as a performance booster.
Excellent surveys of backdoor attack and defense can be found at~\cite{li2020backdoor, liu2020survey, gao2020backdoor, li2020deep}.

\noindent{\bf Interpretation/Attribution.}
\sname{} is also related to model interpretation and attribution, e.g., those identifying important neurons and features~\cite{sundararajan2017axiomatic,shrikumar2016not,ancona2018towards,bau2017network,ancona2017towards,shrikumar2016not,bach2015pixel,das2020massif,dong2017towards,fong2018net2vec,zhou2018interpretable,bau2018identifying,hohman2019s,ghorbani2019towards,yeh2019fidelity}.
The differences lie in that \sname{} focuses on finding distinguishing features of two classes.
Exiting work~\cite{kim2018interpretability} utilizes random examples and training samples of a class (e.g., zebra) to measure the importance of a concept (e.g., `striped') for the class. It however does not find distinguishing internal features.

\section{Conclusion}
We develop a method to distinguish natural and injected backdoors. It is built on a novel symmetric feature differencing technique that identifies a smallest set of features separating two sets of samples. Our results show that the technique is highly effective and enabled us to achieve 
top results on the rounds 2 and 4 leaderboard of the TrojAI competition, and rank the 2nd  in round 3. 
It also shows potential in handling complex and composite semantic-backdoors.

\bibliographystyle{plain}
\bibliography{references}

\clearpage
\appendix

\section{Parameter Settings}
\label{sec:setting}
\sname{} has three hyper-parameters, $\alpha$ to control the weight changes of cross-entropy loss in function (\ref{e:cnmloss}) (in Section~\ref{s:sfd}), $\beta$ to control the similarity comparison between masks in condition (\ref{e:fprule1}) in Section~\ref{s:similarity}
, and $\gamma$ the accuracy threshold in cross-validation checks of masks in Section~\ref{s:similarity}.
We use 0.8, 0.8, and 0.1, respectively, by default.
In our experiments, we use ABS and NC as the upstream scanners. 
The numbers of optimization epochs are 60 for ABS and 1000 for NC. 
The other settings are default unless stated otherwise.

\section{Effects of Hyperparameters}
\label{sec:hyperparameter}
\begin{figure}[]
    \centering                                                   
    \footnotesize 

     \includegraphics[width=0.4\textwidth]{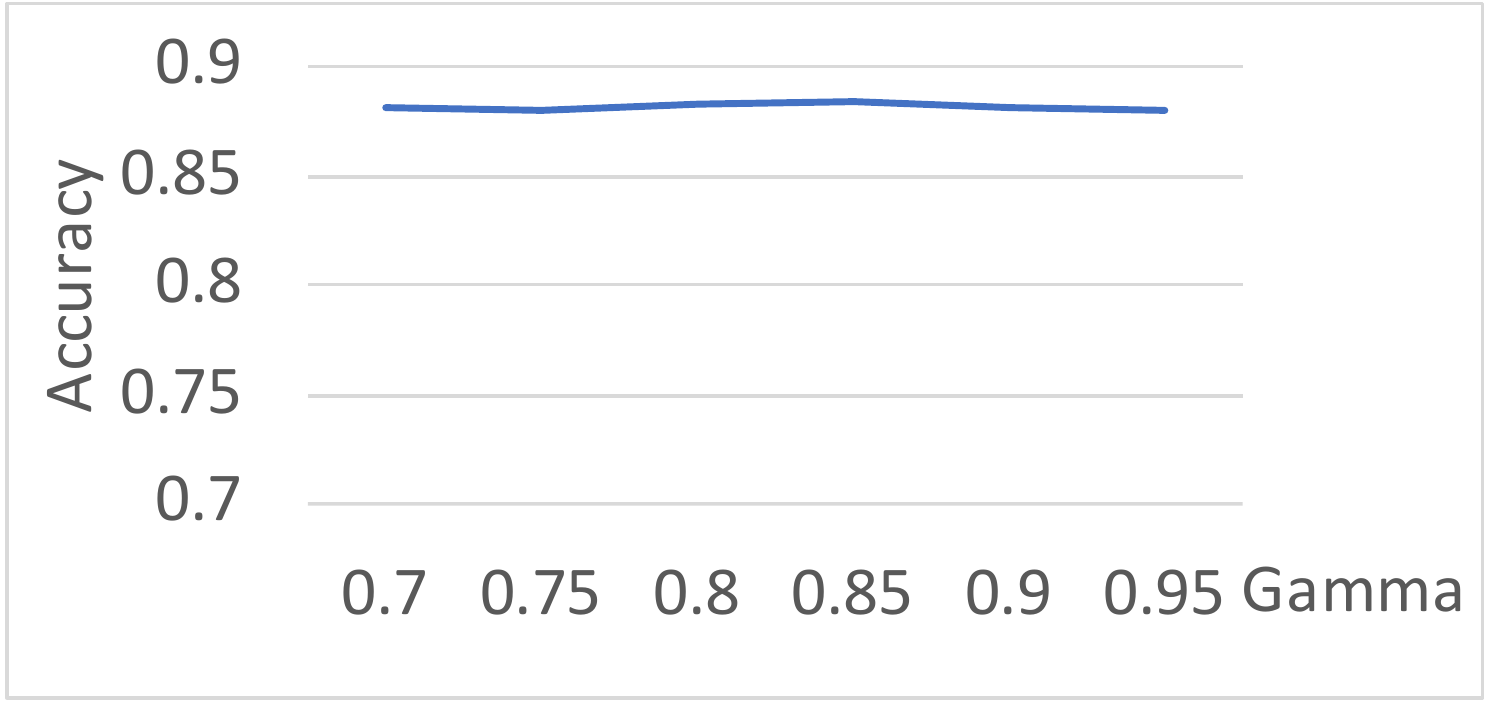}

    \caption{Accuracy changes with $\alpha$ on TrojAI R2 
    } 

   \label{f:acc_alpha}
\end{figure}

\begin{figure}[]
    \centering                                                   
    \footnotesize 

     \includegraphics[width=0.4\textwidth]{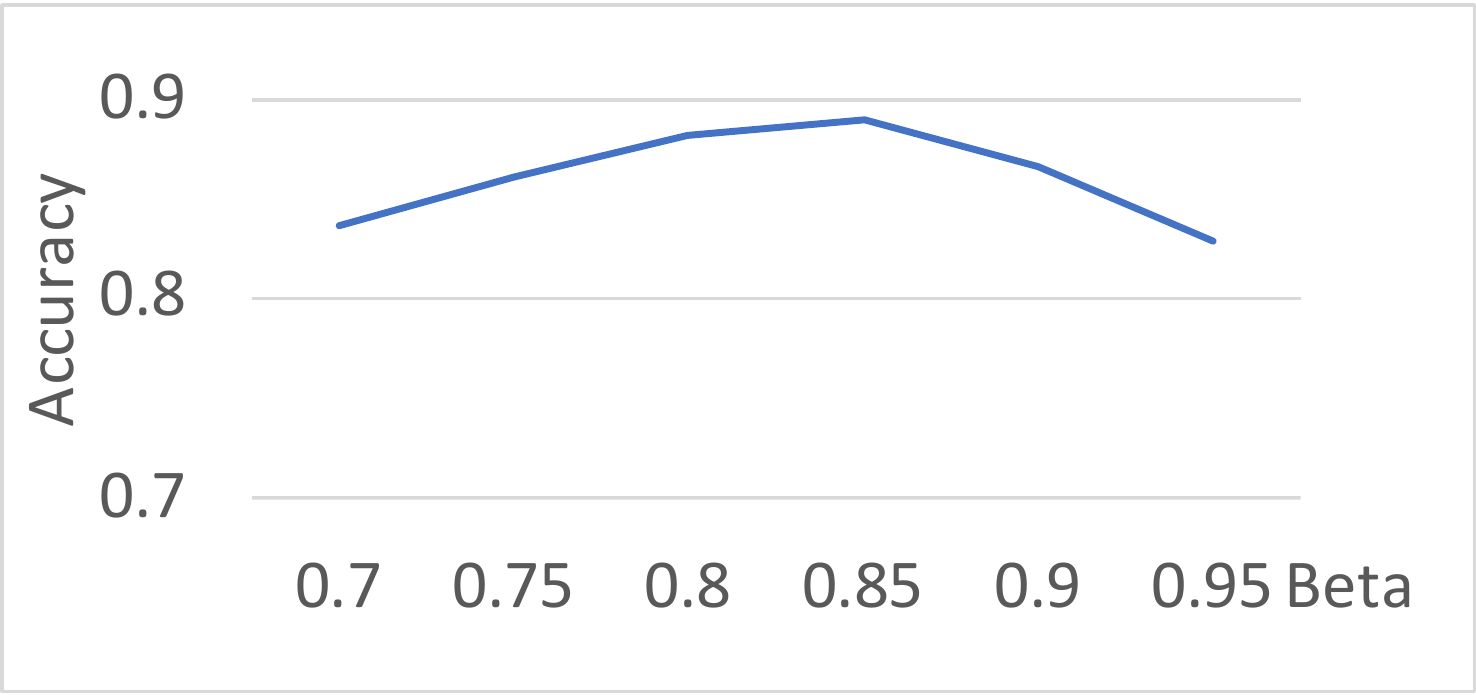}

    \caption{Accuracy changes with $\beta$ on TrojAI R2} 

   \label{f:acc_beta}
\end{figure}

\begin{figure}[]
    \centering                                                   
    \footnotesize 

     \includegraphics[width=0.4\textwidth]{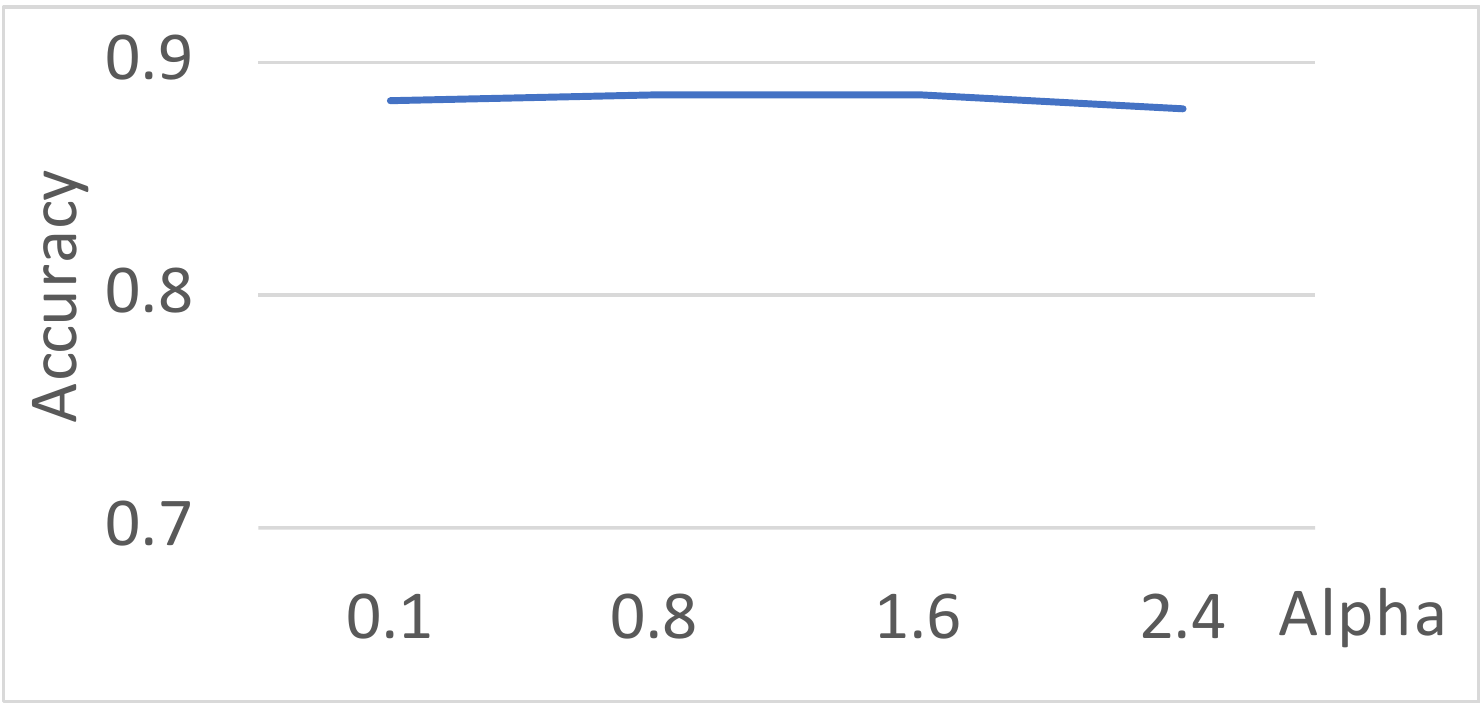}

    \caption{Accuracy changes with $\gamma$ on TrojAI R2} 

   \label{f:acc_gamma}
\end{figure}

We study \sname{} performance with various hyperparameter settings, including the different layer to which \sname{} is applied, different trigger size settings (in the upstream scanner) and different SSIM score bounds (in filter backdoor scanning to ensure the
generated kernel does not over-transform an input), and the 
$\alpha$, $\beta$, and $\gamma$ settings of \sname{}. 
Table~\ref{t:layer} shows the results for layer selection.
The row ``Middle'' means that we apply \sname{} at the layer in the middle of a model. 
The rows ``Last'' and``2nd last'' show the results at the last and the second-last convolutional layers, respectively.
Observe that 
layer selection may affect performance to some extent and the second to the last layer has the best performance. Tables~\ref{t:size} shows that a large trigger size degrades \sname{}'s performance but \sname{} is stable in 900 to 1200. Table~\ref{t:ssim} shows that the SSIM score bound has small effect on performance in  0.7-0.9. Note that an SSIM score smaller than 0.7 means the transformed image is quite different (in human eyes).
Figures~\ref{f:acc_alpha}, ~\ref{f:acc_beta}, and ~\ref{f:acc_gamma} show the performance variations with $\alpha$, $\beta$, and $\gamma$, respectively. The experiments are on the mixture of trojaned models with polygon triggers and the clean models from TrojAI round 2. 
For $\beta$ and $\gamma$, we sample from 0.7 to 0.95 and for $\alpha$ we sample from 0.1 to 2.4. Observe that changing $\alpha$ and $\gamma$ does not have much impact on the overall accuracy.
When we change $\beta$ from 0.7 to 0.95, the overall accuracy is still consistently higher than 0.83. These results show the stability of  \sname{}.

\begin{table}[]
\caption{\sname{} with different trigger sizes; (T:276, C:552) means there are 276 trojaned models and 552 clean models}
\label{t:size}
\centering
\footnotesize
\setlength{\tabcolsep}{5pt}
\begin{tabular}{crrrrrrrrrrr}
\toprule
 & \multicolumn{3}{c}{\begin{tabular}[c]{@{}c@{}}TrojAI R2\\ (T:276,C:552)\end{tabular}} & \multicolumn{1}{c}{} & \multicolumn{3}{c}{\begin{tabular}[c]{@{}c@{}}TrojAI R3\\ (T:252,C:504)\end{tabular}} & \multicolumn{1}{c}{} & \multicolumn{3}{c}{\begin{tabular}[c]{@{}c@{}}TrojAI R4\\ (T:143,C:504)\end{tabular}} \\
 \cmidrule{2-4} \cmidrule{6-8} \cmidrule{10-12}
 & \multicolumn{1}{c}{TP} & \multicolumn{1}{c}{FP} & \multicolumn{1}{c}{Acc} & \multicolumn{1}{c}{} & \multicolumn{1}{c}{TP} & \multicolumn{1}{c}{FP} & \multicolumn{1}{c}{Acc} & \multicolumn{1}{c}{} & \multicolumn{1}{c}{TP} & \multicolumn{1}{c}{FP} & \multicolumn{1}{c}{Acc} \\
900 & 198 & 19 & 0.883 &  & 157 & 19 & 0.849 &  & 95 & 58 & 0.836 \\
1200 & 203 & 30 & 0.876 &  & 175 & 39 & 0.847 &  & 105 & 53 & 0.859 \\
1600 & 210 & 46 & 0.864 &  & 200 & 46 & 0.870 &  & 108 & 77 & 0.827 \\
\bottomrule
\end{tabular}
\end{table}

\begin{table}[]
\caption{\sname{} with different SSIM scores; (T:276, C:552) means there are 276 trojaned models and 552 clean models}
\label{t:ssim}
\footnotesize
\setlength{\tabcolsep}{5pt}
\centering
\begin{tabular}{crrrrrrrrrrr}
\toprule
\begin{tabular}[c]{@{}c@{}}SSIM\\ Score\end{tabular} & \multicolumn{3}{c}{\begin{tabular}[c]{@{}c@{}}TrojAI R2\\ (T:276,C:552)\end{tabular}} & \multicolumn{1}{c}{} & \multicolumn{3}{c}{\begin{tabular}[c]{@{}c@{}}TrojAI R3\\ (T:252,C:504)\end{tabular}} & \multicolumn{1}{c}{} & \multicolumn{3}{c}{\begin{tabular}[c]{@{}c@{}}TrojAI R4\\ (T:361,C:504)\end{tabular}} \\
 \cmidrule{2-4} \cmidrule{6-8} \cmidrule{10-12}
 & \multicolumn{1}{c}{TP} & \multicolumn{1}{c}{FP} & \multicolumn{1}{c}{Acc} & \multicolumn{1}{c}{} & \multicolumn{1}{c}{TP} & \multicolumn{1}{c}{FP} & \multicolumn{1}{c}{Acc} & \multicolumn{1}{c}{} & \multicolumn{1}{c}{TP} & \multicolumn{1}{c}{FP} & \multicolumn{1}{c}{Acc} \\
0.9 & 145 & 4 & 0.837 &  & 115 & 9 & 0.807 &  & 234 & 47 & 0.799 \\
0.8 & 160 & 13 & 0.844 &  & 149 & 39 & 0.812 &  & 242 & 46 & 0.809 \\
0.7 & 204 & 32 & 0.874 &  & 178 & 90 & 0.783 &  & 175 & 13 & 0.770
\\
\bottomrule
\end{tabular}
\end{table}

\begin{table*}[]
\caption{\sname{} with different layer; (T:276,C:552) means that there are 276 trojaned models and 552 clean models}
\label{t:layer}
\footnotesize
\centering
\setlength{\tabcolsep}{5pt}
\begin{tabular}{crrrrrrrrrrrrrrrrrrrrrrr}
\toprule
 & \multicolumn{7}{c}{TrojAI R2} & \multicolumn{1}{c}{} & \multicolumn{7}{c}{TrojAI R3} & \multicolumn{1}{c}{} & \multicolumn{7}{c}{TrojAI R4} \\
 \cmidrule{2-8} \cmidrule{10-16} \cmidrule{18-24} 
 & \multicolumn{3}{c}{\begin{tabular}[c]{@{}c@{}}Polygon Trigger\\ (T:276,C:552)\end{tabular}} & \multicolumn{1}{c}{} & \multicolumn{3}{c}{\begin{tabular}[c]{@{}c@{}}Filter Trigger\\ (T:276,C:552)\end{tabular}} & \multicolumn{1}{c}{} & \multicolumn{3}{c}{\begin{tabular}[c]{@{}c@{}}Polygon Trigger\\ (T:252,C:504)\end{tabular}} & \multicolumn{1}{c}{} & \multicolumn{3}{c}{\begin{tabular}[c]{@{}c@{}}Filter Trigger\\ (T:252,C:504)\end{tabular}} & \multicolumn{1}{c}{} & \multicolumn{3}{c}{\begin{tabular}[c]{@{}c@{}}Polygon trigger\\ (T:143,C:504)\end{tabular}} & \multicolumn{1}{c}{} & \multicolumn{3}{c}{\begin{tabular}[c]{@{}c@{}}Filter trigger\\ (T:361,C:504)\end{tabular}} \\
 \cmidrule{2-4} \cmidrule{6-8} \cmidrule{10-12} \cmidrule{14-16} \cmidrule{18-20} \cmidrule{22-24}  
 & TP & FP & Acc &  & TP & FP & Acc &  & TP & FP & Acc &  & TP & FP & Acc &  & TP & FP & Acc &  & TP & FP & Acc \\
Middle & 215 & 54 & 0.861 &  & 220 & 97 & 0.815 &  & 205 & 100 & 0.806 &  & 153 & 58 & 0.792 &  & 102 & 75 & 0.821 &  & 257 & 77 & 0.791 \\
Second Last & 198 & 19 & 0.883 &  & 204 & 32 & 0.874 &  & 182 & 68 & 0.818 &  & 149 & 39 & 0.812 &  & 105 & 53 & 0.859 &  & 242 & 46 & 0.809 \\
Last & 141 & 6 & 0.83 &  & 171 & 16 & 0.854 &  & 159 & 65 & 0.791 &  & 141 & 27 & 0.817 &  & 84 & 37 & 0.852 &  & 196 & 37 & 0.766 \\
\bottomrule
\end{tabular}
\end{table*}

\section{More Details of the Model Repair Experiment 
}
\label{sec:repair}
We show the trigger size for each label pair for an trojaned model in Table~\ref{t:unlearnexample1} and for an benign model in Table~\ref{t:unlearnexample2}. 
Table~\ref{t:unlearnexample1} (a) shows the trigger size between each pair of labels. The columns denote the victim label and the rows denote the the target label. 
For example, the gray cell in Table~\ref{t:unlearnexample1} (a) shows the trigger size to flip class 1 to class 0. Table~\ref{t:unlearnexample1} (b) follows the same format and shows the result for trojaned model after unlearning. In the trojaned model the injected trigger flips class 1 to class 0. Before unlearning, class 1 and class 0 have the smallest trigger size 21. Unlearning increases the trigger size between the two to 106, which is above the average trigger size between any pairs. Intuitively, one can consider the backdoor is fixed.
In the benign model, the natural trigger flips class 3 to class 5. As shown in Table~\ref{t:unlearnexample2}, unlearning only increases the trigger size from 24 to 59 and 59 is still one of the smallest trigger size among all label pairs for the fixed model.

\begin{figure*}[]
\subfigure[Before unlearning]
{
\begin{minipage}[t]{.45\textwidth}
\footnotesize
\begin{tabular}{|l|l|l|l|l|l|l|l|l|l|l|}
\hline
 & 0 & 1 & 2 & 3 & 4 & 5 & 6 & 7 & 8 & 9 \\ \hline
0 & - & 48 & 34 & 75 & 42 & 52 & 62 & 46 & 48 & 52 \\ \hline
1 & \cellcolor[HTML]{C0C0C0}21 & - & 74 & 91 & 88 & 96 & 72 & 80 & 81 & 45 \\ \hline
2 & 32 & 54 & - & 66 & 39 & 57 & 54 & 61 & 78 & 60 \\ \hline
3 & 34 & 53 & 35 & - & 42 & 27 & 46 & 50 & 72 & 47 \\ \hline
4 & 29 & 45 & 29 & 49 & - & 36 & 46 & 48 & 63 & 48 \\ \hline
5 & 40 & 70 & 35 & 46 & 43 & - & 53 & 49 & 81 & 56 \\ \hline
6 & 29 & 48 & 23 & 41 & 44 & 61 & - & 66 & 70 & 59 \\ \hline
7 & 40 & 77 & 55 & 78 & 40 & 52 & 81 & - & 82 & 60 \\ \hline
8 & 21 & 44 & 42 & 75 & 50 & 59 & 60 & 65 & - & 47 \\ \hline
9 & 29 & 62 & 78 & 85 & 69 & 70 & 73 & 62 & 73 & - \\ \hline
\end{tabular}
\end{minipage}
}
~
\subfigure[After unlearning]
{
\begin{minipage}[t]{.45\textwidth}
\footnotesize
\begin{tabular}{|l|l|l|l|l|l|l|l|l|l|l|}
\hline
 & 0 & 1 & 2 & 3 & 4 & 5 & 6 & 7 & 8 & 9 \\ \hline
0 & - & 84 & 56 & 103 & 77 & 96 & 82 & 96 & 56 & 78 \\ \hline
1 & \cellcolor[HTML]{C0C0C0}106 & - & 132 & 162 & 150 & 140 & 113 & 134 & 124 & 70 \\ \hline
2 & 92 & 111 & - & 88 & 79 & 79 & 62 & 99 & 109 & 106 \\ \hline
3 & 105 & 92 & 66 & - & 86 & 60 & 58 & 82 & 124 & 86 \\ \hline
4 & 96 & 92 & 55 & 81 & - & 72 & 53 & 77 & 99 & 95 \\ \hline
5 & 119 & 100 & 64 & 70 & 91 & - & 58 & 101 & 136 & 90 \\ \hline
6 & 107 & 97 & 86 & 88 & 99 & 93 & - & 113 & 113 & 97 \\ \hline
7 & 94 & 101 & 92 & 124 & 87 & 87 & 81 & - & 126 & 100 \\ \hline
8 & 50 & 72 & 68 & 104 & 98 & 106 & 81 & 101 & - & 79 \\ \hline
9 & 104 & 87 & 129 & 119 & 117 & 115 & 110 & 108 & 123 & - \\ \hline
\end{tabular}
\end{minipage}
}
\caption{Injected trigger distance matrix before and after unlearning}
\label{t:unlearnexample1}
\end{figure*}

\begin{figure*}[]
\subfigure[Before unlearning]
{
\begin{minipage}[t]{.45\textwidth}
\footnotesize
\begin{tabular}{|l|l|l|l|l|l|l|l|l|l|l|}
\hline
 & 0 & 1 & 2 & 3 & 4 & 5 & 6 & 7 & 8 & 9 \\ \hline
0 & - & 43 & 44 & 47 & 38 & 42 & 67 & 68 & 37 & 48 \\ \hline
1 & 62 & - & 89 & 88 & 79 & 78 & 70 & 85 & 76 & 47 \\ \hline
2 & 53 & 58 & - & 37 & 35 & 42 & 51 & 71 & 72 & 62 \\ \hline
3 & 62 & 66 & 40 & - & 40 & \cellcolor[HTML]{C0C0C0}24 & 48 & 59 & 72 & 56 \\ \hline
4 & 61 & 57 & 38 & 48 & - & 31 & 52 & 52 & 85 & 64 \\ \hline
5 & 69 & 66 & 43 & 33 & 46 & - & 51 & 61 & 73 & 57 \\ \hline
6 & 66 & 55 & 32 & 35 & 44 & 38 & - & 80 & 87 & 62 \\ \hline
7 & 74 & 77 & 67 & 61 & 38 & 39 & 74 & - & 92 & 68 \\ \hline
8 & 29 & 44 & 55 & 61 & 48 & 51 & 62 & 65 & - & 46 \\ \hline
9 & 79 & 57 & 84 & 72 & 67 & 67 & 83 & 76 & 72 & - \\ \hline
\end{tabular}
\end{minipage}
}
~
\subfigure[After unlearning]
{
\begin{minipage}[t]{.45\textwidth}
\footnotesize
\begin{tabular}{|l|l|l|l|l|l|l|l|l|l|l|}
\hline
 & 0 & 1 & 2 & 3 & 4 & 5 & 6 & 7 & 8 & 9 \\ \hline
0 & - & 79 & 56 & 89 & 63 & 90 & 65 & 99 & 59 & 76 \\ \hline
1 & 120 & - & 134 & 114 & 123 & 154 & 77 & 119 & 122 & 48 \\ \hline
2 & 95 & 101 & - & 75 & 58 & 74 & 59 & 82 & 121 & 82 \\ \hline
3 & 104 & 98 & 57 & - & 58 & \cellcolor[HTML]{C0C0C0}59 & 40 & 90 & 124 & 80 \\ \hline
4 & 104 & 100 & 60 & 88 & - & 79 & 50 & 73 & 117 & 79 \\ \hline
5 & 95 & 91 & 58 & 84 & 76 & - & 52 & 77 & 131 & 86 \\ \hline
6 & 114 & 115 & 77 & 129 & 102 & 89 & - & 131 & 137 & 93 \\ \hline
7 & 104 & 120 & 110 & 103 & 68 & 92 & 66 & - & 129 & 78 \\ \hline
8 & 36 & 61 & 74 & 80 & 66 & 112 & 66 & 86 & - & 70 \\ \hline
9 & 128 & 109 & 117 & 103 & 112 & 120 & 73 & 124 & 105 & - \\ \hline
\end{tabular}
\end{minipage}
}
\caption{Natural trigger distance matrix before and after unlearning}
\label{t:unlearnexample2}
\end{figure*}

\end{document}